\documentclass[pdflatex,sn-mathphys-num]{sn-jnl}

\usepackage{graphicx}%
\usepackage{multirow}%
\usepackage{amsmath,amssymb,amsfonts}%
\usepackage{amsthm}%
\usepackage{mathrsfs}%
\usepackage[title]{appendix}%
\usepackage{xcolor}%
\usepackage{textcomp}%
\usepackage{manyfoot}%
\usepackage{booktabs}%
\usepackage{algorithm}%
\usepackage{algorithmicx}%
\usepackage{algpseudocode}%
\usepackage{listings}%
\usepackage{comment}

\usepackage{float}

\usepackage{longtable} %
\usepackage{microtype}
\usepackage{pdflscape}

\definecolor{darkgreen}{rgb}{0.0, 0.55, 0.0}
\definecolor{darkblue}{rgb}{0.0, 0.0, 0.55}

\theoremstyle{thmstyleone}%
\theoremstyle{thmstyletwo}%

\theoremstyle{thmstylethree}%

\begin{document}


\title[Article Title]{The Shift Toward Open and Reproducible AI Research}


\author*[1,2]{\fnm{Kevin L} \sur{Coakley}}\email{kcoakley@sdsc.edu}

\author[3]{\fnm{Thijs} \sur{Snelleman}}\email{snelleman@aim.rwth-aachen.de}

\author[3,4]{\fnm{Holger} \sur{Hoos}}\email{hh@aim.rwth-aachen.de}

\author*[1]{\fnm{Odd Erik} \sur{Gundersen}}\email{odderik@ntnu.no}

\affil*[1]{\orgdiv{Department of Computer Science}, \orgname{Norwegian University of Science and Technology}, \orgaddress{\street{Sem Sælands Vei 7-9}, \city{Trondheim}, \postcode{7031}, \state{Trønderlag}, \country{Norway}}}

\affil[2]{\orgdiv{San Diego Supercomputer Center}, \orgname{University of California San Diego}, \orgaddress{\street{9500 Gilman Dr}, \city{La Jolla}, \postcode{92093}, \state{California}, \country{USA}}}

\affil[3]{\orgdiv{Chair of AI Methodology}, \orgname{RWTH Aachen University}, \orgaddress{\street{Theaterstrasse 35-39}, \city{Aachen}, \postcode{52062}, \state{North Rhine-Westphalia}, \country{Germany}}}

\affil[4]{\orgdiv{Leiden Institute of Advanced Computer Science}, \orgname{Leiden University}, \orgaddress{\street{Einsteinweg 55}, \city{Leiden}, \postcode{2333 CC}, \state{South Holland}, \country{The Netherlands}}}


\abstract{
    The reproducibility crisis has directed the AI research community toward improving documentation practices.
    Several studies have identified methodological issues, and in response, the most impactful venues in the field have introduced reproducibility checklists.
    We seek to understand whether documentation practices have changed over time by assessing all published papers at five leading AI conferences over the past decade. 
    Seven reproducibility variables were identified, quality-assured and used to analyse 56\,800 publications.
    Our analysis reveals that in the period 2014 to 2024, documentation practices have improved; papers sharing both code and data increased nearly sixfold, from 11\% to 64\%
    Building on empirical reproducibility rates from a prior study, we estimate — inferred from documentation practices, not direct testing — that reproducibility increased from 28\% in 2014 to 64\% in 2024.
    Improvements in documentation practices predate the introduction of reproducibility checklists, suggesting these changes reflect a broader movement toward open science rather than a direct response to formal requirements.
}

\maketitle

Many scientific results cannot be reliably reproduced~\cite{ioannidis_2005}, and therefore conclusions cannot be trusted \cite{mcnutt_2014}. 
This phenomenon is called the reproducibility crisis~\cite{baker_2016} and has been described in a number of studies covering a broad range of disciplines, including psychology~\cite{pashler_2012, open_2015, klein_2014}, economics~\cite{camerer_2016}, medicine~\cite{prinz_2011, collins_2014, begley_2012}, neuroscience~\cite{button_2013}, and genetics~\cite{hewitt_2012}. 
Reproducibility concerns also gained prominence in artificial intelligence (AI)~\cite{gundersen_2018, hutson2018reproduciblitycrisis}, where rapid growth in the field and increasing social importance have intensified the urgency to mitigate the problem.

\begin{figure}[!t]
    \centering
    \textbf{Annual publication counts across five leading AI conferences from 2014 to 2024.}\\[0.5em]
    \includegraphics[width=1\linewidth]{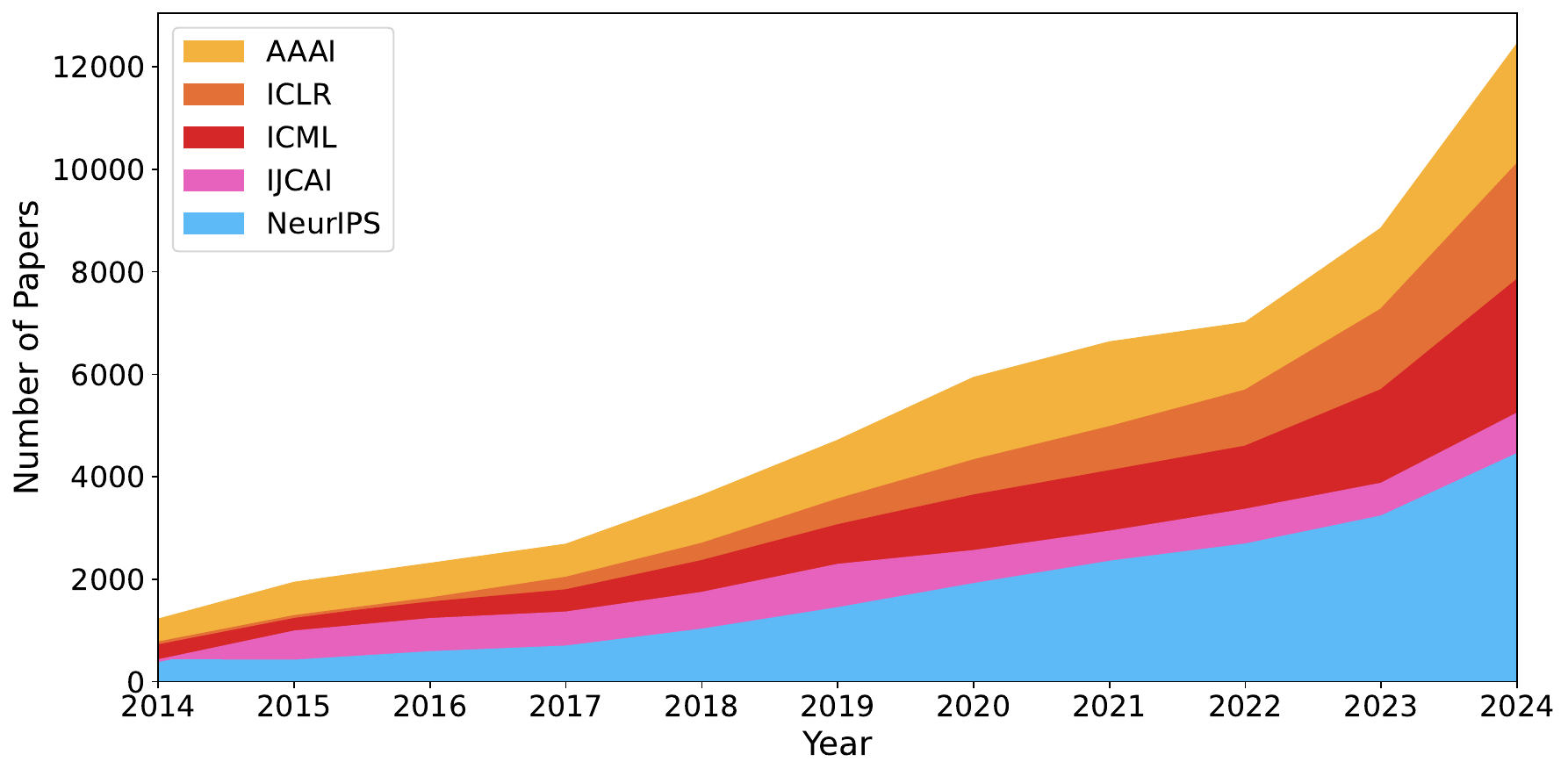}
    \caption{The number of papers published at five leading AI conferences (AAAI, ICLR, ICML, IJCAI, and NeurIPS) from 2014 to 2024. Publications increased tenfold over this period, from 1\,206 papers in 2014 to 12\,026 in 2024, reflecting the rapid expansion of AI research. IJCAI did not hold a conference in 2014. A tabular representation of this data can be found in Supplementary Table 10.}
    \label{figure:papers_per_conference}
\end{figure}

One way to mitigate the reproducibility crisis is to provide detailed descriptions of the research alongside open code and data, and therefore a cultural shift toward open science~\cite{vicente_2018} has emerged, advocating for transparent and accessible knowledge.
As a large portion of empirical research in computer science and AI is conducted on computers and experiments are described in code, sharing detailed descriptions of experiments should, in principle, be straightforward~\cite{bischl2025openml}.
Achieving this relies critically on high-quality documentation of experimental methods and artifacts, such as code, data, and workflows~\cite{foster2017osf,wilkinson2016fair,white2024model,wilkinson2025applying}.
This has again led communities and venues to demand stronger requirements from authors, reflected in the introduction of reproducibility checklists for example at NeurIPS~\cite{pineau_2021} and JAIR~\cite{gundersen2024improving}.
However, the effect of these procedural and communal changes remains difficult to measure, especially at scale, due to the manual labour required to reproduce studies  
and the tenfold increase in published studies in the past decade, see \autoref{figure:papers_per_conference}.

It is widely acknowledged that empirical AI research can fail in subtle ways. 
Nondeterminism due to variability between execution (or computing) environments can cause results to vary so much that it affects reproducibility~\cite{lucic_2018, henderson_2018, ferrari_2019, belz_2022, gundersen_2023}.
Poor data treatment, including poorly documented datasets, dataset splits, and data leakage also 
pose challenges to reproducibility~\cite{gebru_2021, varoquaux_2022, kapoor_2023}. 
Inadequate hyperparameter optimisation can lead to irreproducible results~\cite{haibe_2020, belz_2021, lucic_2018, bouthillier_2019}, and the same holds for parallelization~\cite{hunold2016reproducible}, compiler settings, and operating system~\cite{hong2013evaluation}.
Even when code is provided, software dependency issues, errors, and post-publication updates can prevent empirical AI research from being reproduced~\cite{stodden_2016,ajayi_2023}.
\citet{gundersen_2022} categorizes different sources of irreproducibility, but few papers investigate to what degree empirical AI research results can be trusted. 

Several attempts have been made to quantify the reproducibility of empirical research in computer science and AI.
Following ~\citet{gundersen_2021}, we use reproducibility to refer to the ability of independent investigators to draw the same conclusions from an experiment by following the documentation shared by the original investigators.
No distinction is made between reproducibility and replication, consistent with \citet{gundersen_2021}, \citet{schmidt_2009}, \citet{nosek_2014}, \citet{goodman_2016}, and \citet{gundersen_2018}.
~\citet{collberg_2016} were only able to execute the code of 48.3\% of the 601 computer science papers that shared code as part of the publication, not even attempting to evaluate whether the results were reproduced. 
\citet{raff_2019} was able to reproduce 63.4\% of 255 papers by reimplementing algorithms from scratch, while \citet{gundersen_2025} reproduced 33\% of papers that only provided data and 86\% of papers that shared both code and data.

These studies, while insightful, reveal intrinsic limitations of manual reproduction. 
Efforts are limited by the time required for manual reproduction, 
which introduces selection biases when choosing which papers to replicate and reduces 
sample size.
Accurately measuring reproducibility trends in a field in which the leading five conferences publish more than 12 \,000 papers annually cannot be done manually.

Another approach avoids reimplementing experiments entirely and instead focuses on the quality of documentation and the inclusion of open artifacts as a practical means of evaluating reproducibility in the spirit of open science.  
\citet{gundersen_2018} manually evaluated 400 randomly selected papers for characteristics associated with reproducibility. 
Random selection reduced selection bias and documentation analysis allowed for increasing the sample size to 400, yet their results still represent only a fraction of conference papers published between 2013 and 2016.
The reliance on manual effort, whether for full reproduction or documentation analysis, renders a large-scale, longitudinal analysis of the impact of open science on empirical AI research infeasible.

To address this methodological gap and, to the best of our knowledge, provide the first comprehensive longitudinal analysis of open science in AI research, in this study, we automate the evaluation of reproducibility documentation at scale leveraging large language models (LLMs). 
We systematically assess documentation practices and the use of open artifacts, such as code and datasets, of 56\,800 papers from five leading AI conferences — AAAI (Association for the Advancement of Artificial Intelligence Conference on Artificial Intelligence), ICLR (International Conference on Learning Representations), ICML (International Conference on Machine Learning), IJCAI (International Joint Conference on Artificial Intelligence), and NeurIPS (Conference on Neural Information Processing Systems) — spanning from 2014 to 2024.
We confirmed the reliability of our approach through prompt optimisation against a manually annotated dataset; furthermore, we manually evaluating 160 randomly sampled papers from the resulting dataset.
This automated method circumvents the prohibitive costs and inherent selection biases of manual reproduction, as well as the costs of manual documentation analysis that have constrained previous work, enabling an, previously infeasible, exhaustive analysis. 
With this scalable framework, our study tracks trends in artifact sharing, quantifies the impact of community practices, such as requirements for reproducibility checklists, and characterizes the evolution of the estimated reproducibility rate of AI research.

Our analysis reveals a quantifiable cultural shift toward open science through improved quality of documentation and increased artifact sharing within the AI research community.
We estimate that the reproducibility rate of empirical research — inferred from documentation practices and the empirical rates reported by \citet{gundersen_2025}, not direct testing — has more than doubled, increasing from 28\% in 2014 to 64\% in 2024, a trend supported by a fivefold increase in papers sharing both code and relying on open datasets.
Our results show that improvements in documentation practices were underway before reproducibility checklists were introduced.
Our study advances the literature on AI reproducibility by demonstrating quantifiable trends in documentation practices, validated against manually annotated datasets, while establishing an LLM-based methodology for conducting meta-science at scale.

\section{Results}

We analysed 56\,800 papers from five top-tier AI and ML conferences, AAAI, ICLR, ICML, IJCAI, and NeurIPS, published between 2014 and 2024. 
The number of publications at these conferences has grown substantially over this period, from 1\,206 papers in 2014 to 12\,026 in 2024, a tenfold increase.
52\,328 papers (92\%) were identified as empirical research and thus form the basis of our analysis; we excluded theoretical papers, as the reproducibility of theoretical work warrants an analysis of an entirely different nature.

\subsection{Reproducibility Variables}
\label{sec:reproducibility-variables}

To quantify the progress of open science and documentation practices in empirical AI research, we evaluated each paper in our dataset for a set of variables. 
These variables have been proposed in other studies \cite{gundersen_2018,raff_2019,magnusson_2023} and cover the previously introduced reproducibility checklists (see Supplementary Table 1). 
We use the \emph{reproducibility variables} as seven Boolean indicators that represent whether a given paper contains the following information:
(1) availability of open source code;
(2) use of one or more open datasets; 
(3) specification of dataset splits; 
(4) pseudocode describing the algorithm(s) used;
descriptions of (5) the hardware used to conduct the experiment; 
(6) the software dependencies; 
and (7) the setup of the experiment. 
In addition, we distinguish whether the study is considered empirical or theoretical, and capture whether at least one author has an industry (rather than academic) affiliation to assess how institutional context influences documentation practices.

\subsection{Improved Documentation and Open Science Practices}
\label{sec:improved-documentation-and-open-science-practices}

\begin{figure}[!t]
    \centering
    \textbf{Distribution of empirical papers by number of documented reproducibility variables from 2014 to 2024.}\\[0.5em]
    \includegraphics[width=1\linewidth]{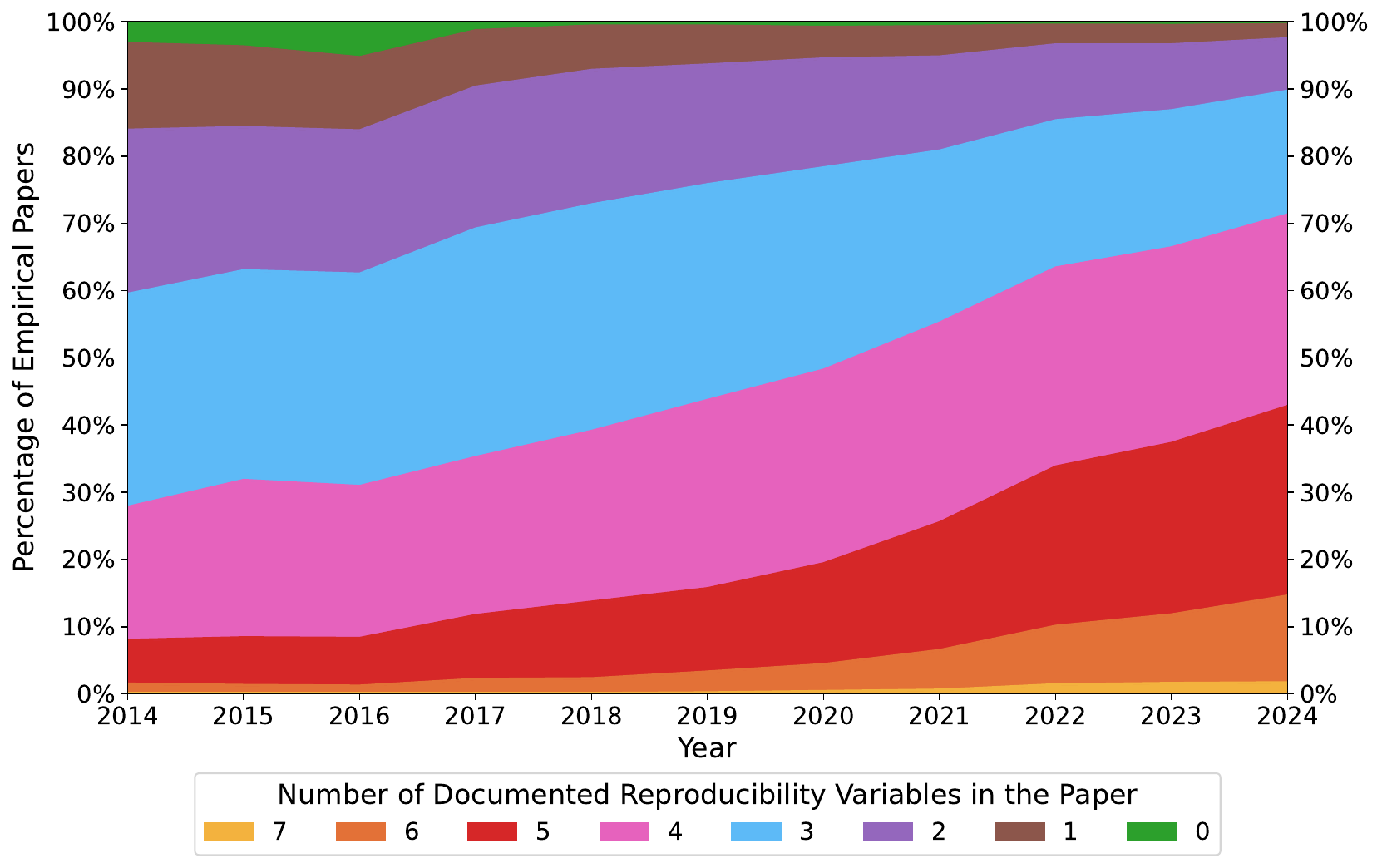}
    \caption{The distribution of empirical papers grouped by how many of the seven reproducibility variables (pseudocode, open code, open datasets, dataset splits, hardware specification, software dependencies, and experiment setup) they document, shown for each year from 2014 to 2024. Each coloured band represents papers documenting a specific count of variables, from zero (top, dark green) to all seven (bottom, light orange). Documentation comprehensiveness improved substantially: the proportion of papers documenting none of these variables decreased from 3\% to 0.3\%, while papers documenting all seven increased from 0.1\% to 1.7\%. Most notably, papers documenting at least five of the seven variables increased more than fivefold, from 8\% in 2014 to 43\% in 2024, demonstrating a sustained shift toward more thorough  documentation practices.}    
    \label{figure:num_reproducibility_variables}
\end{figure}

Research documentation has improved from 2014 to 2024; we observed a substantial positive trend for reproducibility variables.
In 2014, only 8\% of the empirical papers provided documentation for five or more reproducibility variables (\autoref{figure:num_reproducibility_variables}), while by 2024 this figure had increased to 43\%, representing more than a fivefold increase over the decade.
The fraction of papers providing documentation for none of these variables vanished by 2019; at the same time, papers providing documentation for all variables emerged.

The seven reproducibility variables were selected because each captures a distinct artifact or methodological description that reduces ambiguity for an independent team attempting to reproduce an experiment. 
Because empirical AI research encompasses a wide variety of methodologies, from reinforcement learning in simulated environments to benchmarking on static corpora, none of the seven is strictly required for every paper, and documenting all seven does not guarantee reproducibility. 
However, the absence of each variable introduces a specific barrier: without pseudocode or open code, algorithmic ambiguity increases~\cite{gundersen_2018_2,raff_2019}; without open datasets or dataset splits, an independent team must invest substantial time and compute identifying data and splits that yield comparable results~\cite{gundersen_2025,makridakis_2018,pouchard_2020}; without hardware and software specifications, environmental differences can shift results enough to alter conclusions, particularly when performance margins are small~\cite{coakley_2022,hong2013evaluation,zhuang_2021}; and without experiment setup details, hyperparameter search requires additional time and compute~\cite{cooper_2021,raff_2019,reimers_2017}.
Each documented variable therefore removes one such barrier. 
At the population level, papers that document more variables leave fewer barriers in place, increasing the likelihood of successful reproduction. 
The year-over-year increase in the proportion of papers documenting five or more variables reflects a measurable shift in the expected reproducibility of the published literature, not a guarantee for any individual paper.

\begin{figure}[!t]
    \centering
    \textbf{Open code and open dataset documentation rates across five AI conferences from 2014 to 2024.}\\[0.5em]
    \includegraphics[width=1\linewidth]{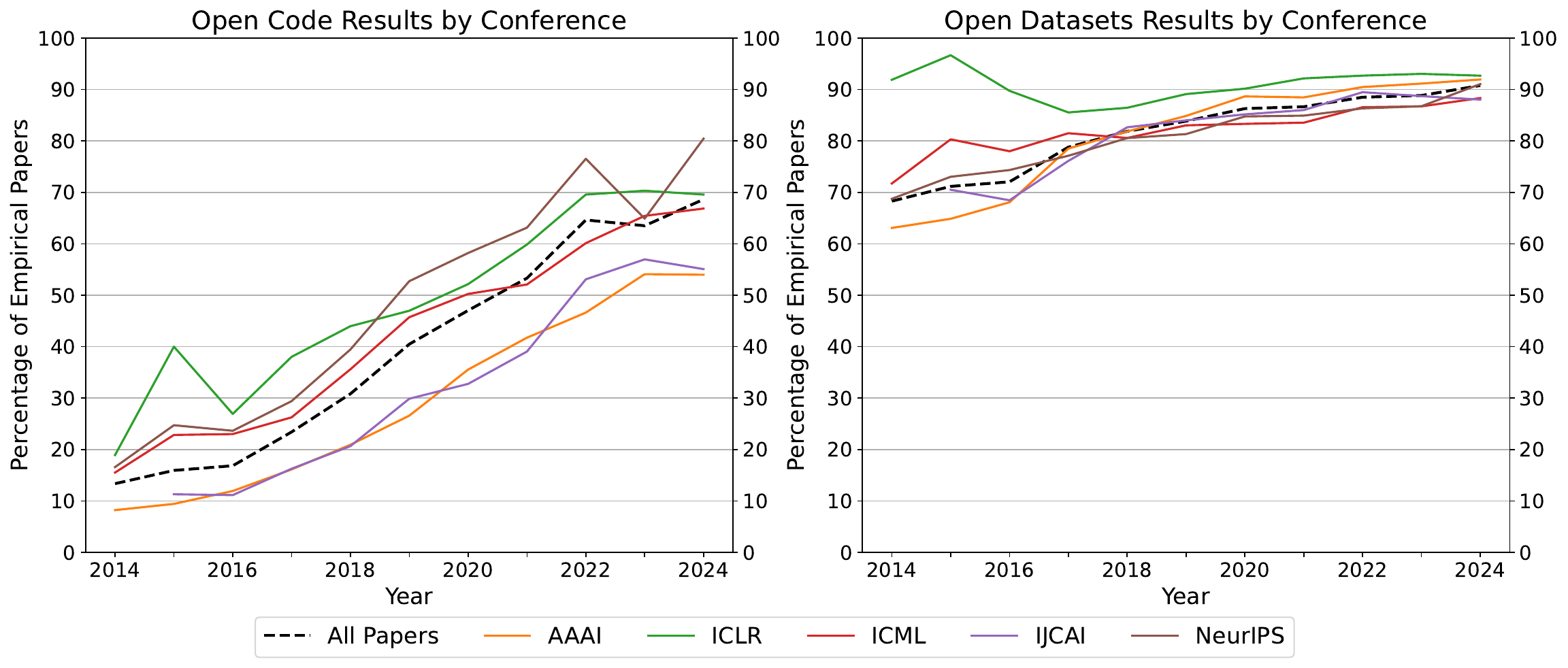}
    \caption{The percentage of empirical papers documenting the availability of open code (left) and the use of open datasets (right) for each of the five conferences from 2014 to 2024. IJCAI did not hold a conference in 2014. Both metrics show substantial increases across all venues. Open code availability rose from 13\% in 2014 to 69\% in 2024 across all conferences (dashed black line), while open dataset usage increased from 68\% to 91\% over the same period.}
    \label{figure:open_code_data_conference}
\end{figure}

\begin{figure}[!t]
    \centering
    \textbf{Rates of empirical papers documenting both open code and open datasets versus neither from 2014 to 2024.}\\[0.5em]
    \includegraphics[width=1\linewidth]{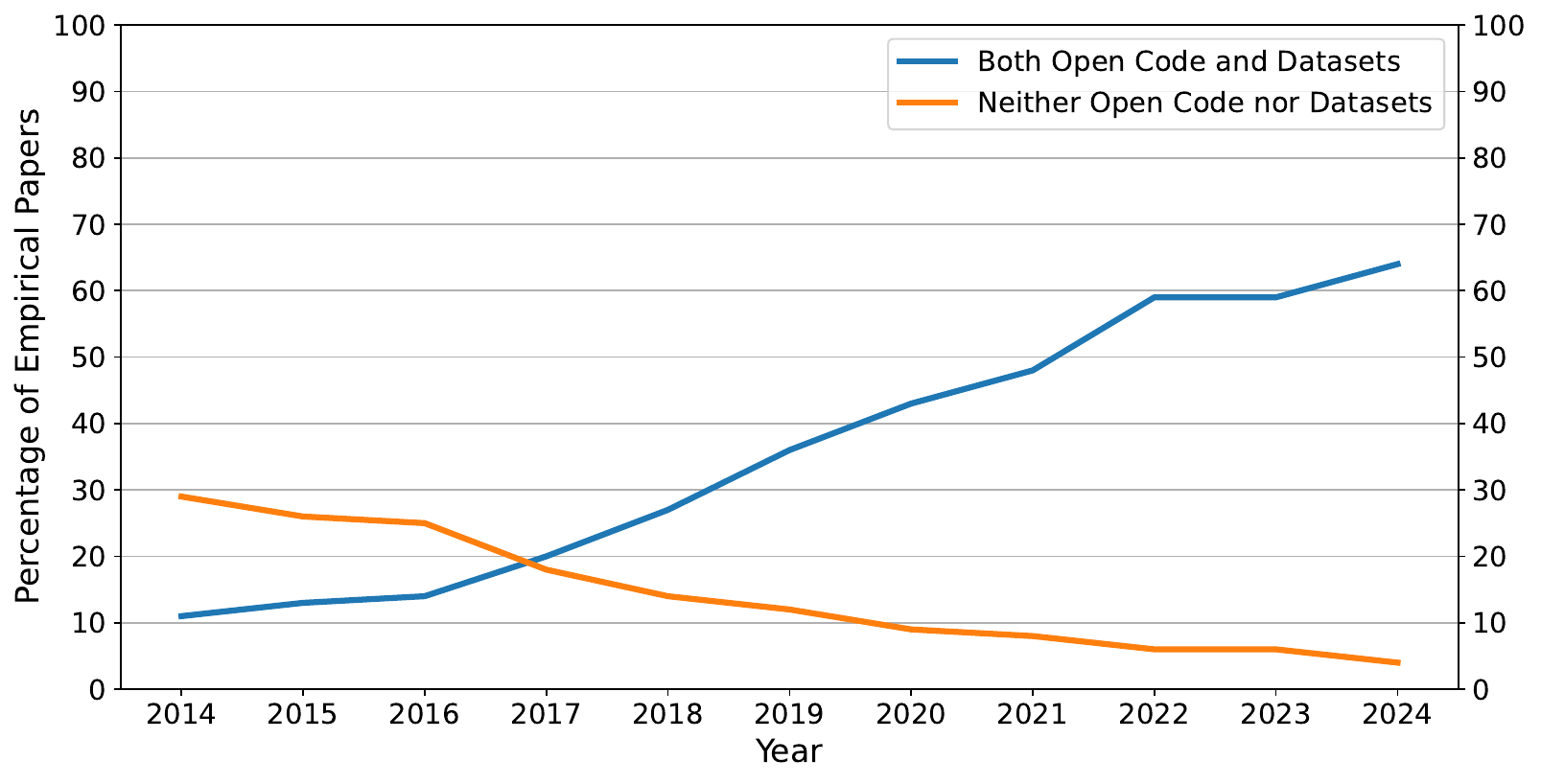}
    \caption{The percentage of all empirical papers across the five conferences that document both the availability of open code and the use of open datasets (blue) versus papers that document neither (orange), from 2014 to 2024. Papers sharing both code and data increased nearly sixfold, from 11\% in 2014 to 64\% in 2024, while papers sharing neither decreased sevenfold, from 29\% to 4\% over the same period. This shift represents a fundamental change in research transparency, as the combination of open code and open datasets is most strongly associated with reproducible findings~\cite{gundersen_2025}. The crossing point occurred around 2017, after which sharing both became more common than sharing neither.}
    \label{figure:both_open_code_data}
\end{figure}

Our analysis shows that the AI community has broadly adopted open science practices.
Open science promotes transparency, accessibility and reproducibility in research, and sharing of code and data is a cornerstone for achieving this~\cite{bischl2025openml}.
Our results show an increasing trend towards open source code and open datasets, as seen in~\autoref{figure:open_code_data_conference}; the use of open datasets increased by 23 percentage points from 68\% to a of 91\%, whereas open source code increased five-fold from 13\% to 69\%.

Sharing code and data correlates with successful replication~\cite{gundersen_2025}.
Over the time period we analysed, the number of papers that provide open source code and open data, the combination most strongly associated with reproducible findings, increased fivefold, while the proportion that shares neither code nor data, which requires the most effort to reproduce, decreased by a factor of seven, as illustrated in~\autoref{figure:both_open_code_data}.
The substantial growth in papers sharing both code and data is associated with a higher probability of reproducible findings.

\begin{figure}[!t]
    \centering
    \textbf{Estimated reproducibility rate of empirical AI papers from 2014 to 2024.}\\[0.5em]
    \includegraphics[width=1\linewidth]{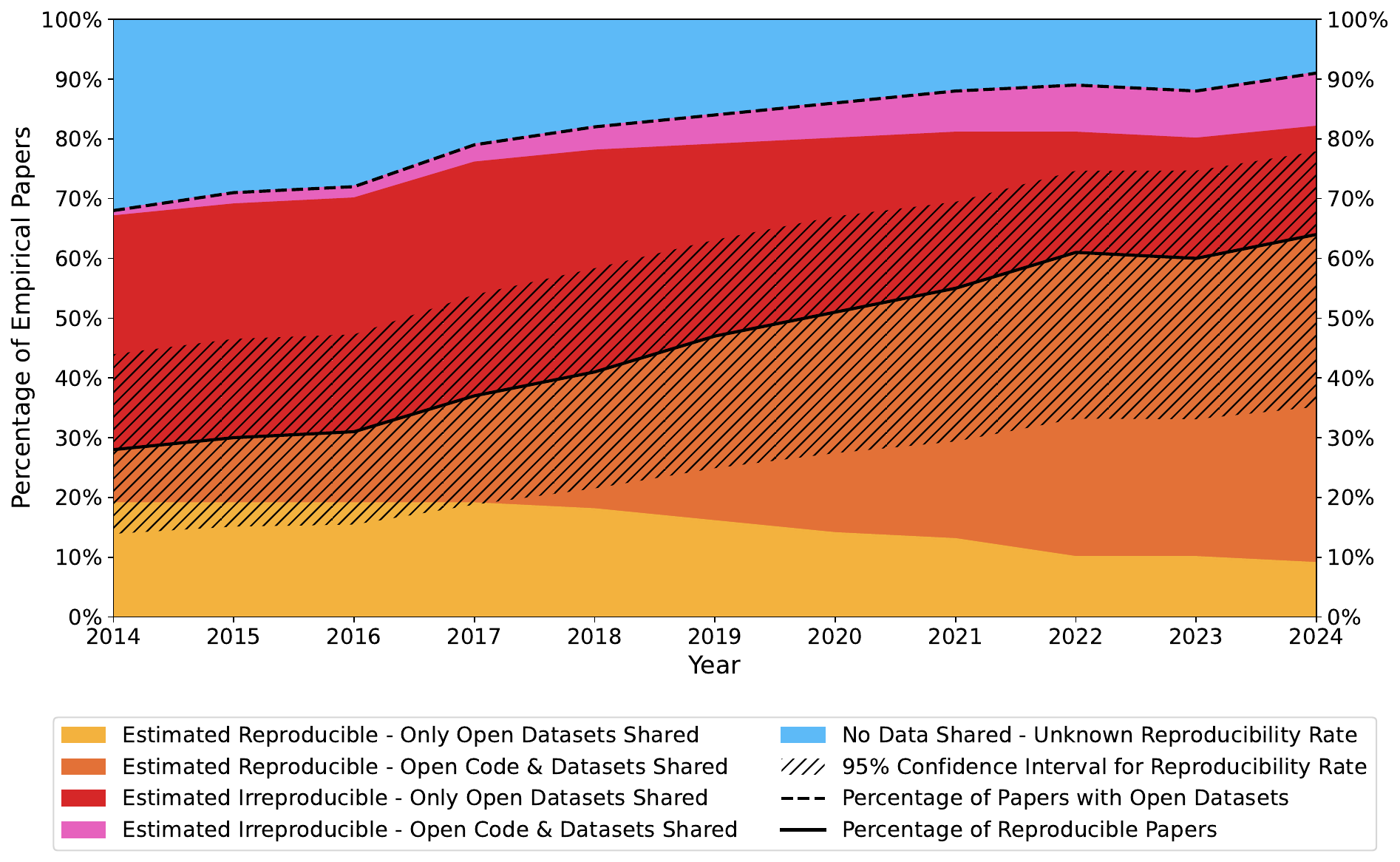}
    \caption{Estimated reproducibility rate of empirical AI research from 2014 to 2024. The estimated reproducibility of papers based on the availability of open code and open datasets, applying the empirical reproducibility rates reported by \citet{gundersen_2025}: 33\% for papers using open datasets only and 86\% for papers sharing both code and open datasets. The overall estimated reproducibility rate (dot-dash black line) more than doubled from 28\% in 2014 to 64\% in 2024. The dashed black line shows the percentage of papers using open datasets, which directly determines whether reproducibility can be estimated, increased from 68\% to 91\% of all empirical papers, while papers with unknown reproducibility (blue, no open datasets) decreased from 32\% to 9\%. The hatched band represents the 95\% confidence interval for the estimated reproducibility rate, derived from a Monte Carlo simulation using Beta distributions parametrised from the replication counts of \citet{gundersen_2025}.} This substantial improvement demonstrates the tangible impact of increased code and data sharing on the reproducibility of AI research.
    \label{figure:reproducibility_rate}
\end{figure}

To estimate what proportion of publications can be reproduced, we performed an analysis based on the empirical reproducibility rates reported by \citet{gundersen_2025}. 
Specifically, we applied two fixed rates from that study: 33\% for papers sharing only open datasets, and 86\% for papers sharing both open code and open datasets. 
For each year, we computed a weighted sum of these rates using the observed proportions of papers in each category as weights, providing the estimated reproducibility rate for each year (\autoref{figure:reproducibility_rate}). 
This procedure assumes that the reproducibility rates from \citet{gundersen_2025} hold as constants across all five conferences and the full 2014–2024 period, and that papers using private datasets can be excluded from the estimate; the resulting values should therefore be interpreted as approximations rather than precise measurements.

To quantify the uncertainty in these estimates, we computed Wilson score 95\% confidence intervals for the two empirical reproducibility rates reported by \citet{gundersen_2025}: 86\% (6/7) for papers sharing both open code and open datasets and 33\% (5/15) for papers sharing only open datasets. 
The lower and upper bounds were then applied to the per-year reproducibility rate estimate (\autoref{figure:reproducibility_rate}).

Over the 11 years covered by our study, our results show a substantial increase in the estimated reproducibility rate, inferred from the documentation of the reproducibility variables, not direct testing, more than doubled from 28\% in 2014 to 64\% in 2024 (\autoref{figure:reproducibility_rate}).
The proportion of empirical papers using private datasets, for which reproducibility could not be estimated, decreased by 72\%, from 32\% of the research analysed in 2014 to 9\% in 2024. 
Although the overall number of published empirical papers increased tenfold during that time, the number of papers estimated to be reproducible increased nearly 24-fold, from approximately 305 in 2014 to approximately 7265 in 2024.
These results illustrate how increased documentation, especially in the form of open code and  datasets, and the broader adoption of open science practices have tangibly improved the rigour of AI research, even as the field has expanded rapidly.

\subsection{The Effect of Introducing Reproducibility Checklists}

\begin{figure}[!t]
    \centering
    \textbf{Reproducibility variable documentation rates with pre- and post-checklist trend lines, across all five conferences from 2014 to 2024.}\\[0.5em]
    \includegraphics[width=.95\linewidth]{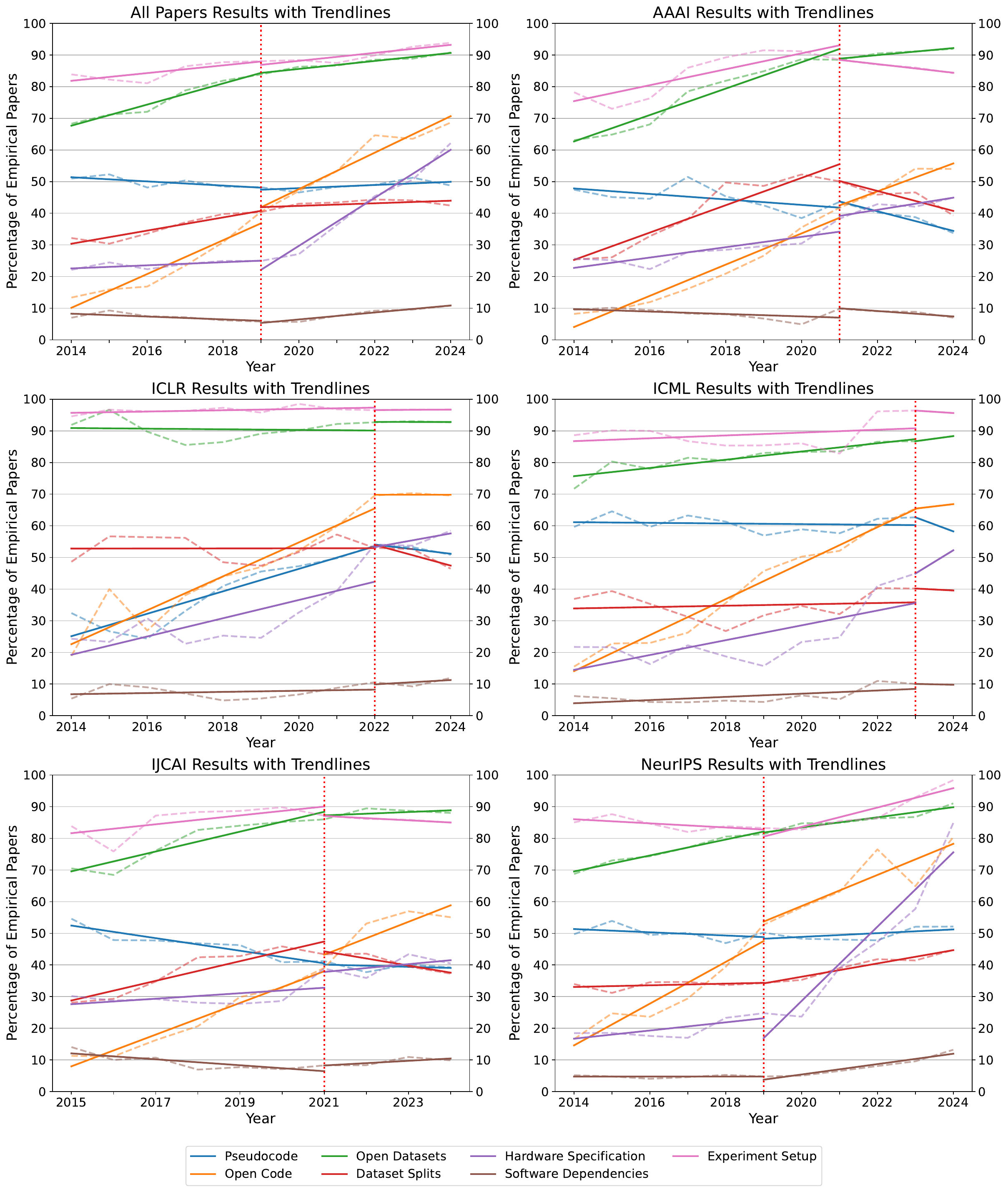}
    \caption{The percentage of empirical papers documenting seven reproducibility variables across all five conferences and for each conference individually from 2014 to 2024. Dashed lines show the observed annual percentages for each reproducibility variable; solid lines show the trend lines fitted before and after reproducibility checklist introduction, with slopes quantifying the rate of change in percentage points per year (pp/year). For aggregated results across all conferences, 2019 serves as the reference year, corresponding to NeurIPS's introduction of the first reproducibility checklist. IJCAI did not hold a conference in 2014. The trends demonstrate that improvements in documentation practices were largely established before formal checklist requirements, with the notable exception of hardware specification, which showed increased slopes after checklist introduction, particularly at NeurIPS.}
    \label{figure:trends_all_conferences}
\end{figure}

\begin{table}[h!]
\centering
\textbf{Reproducibility variable trend slopes (pp/year) before and after checklist introduction, by conference.}\\[0.5em]
\begin{tabular}{l*{3}{rr}}
\toprule
\multirow{2}{*}{Reproducibility Variable} & \multicolumn{2}{c}{All Papers} & \multicolumn{2}{c}{AAAI} & \multicolumn{2}{c}{ICLR} \\
\cmidrule(lr){2-3} \cmidrule(lr){4-5} \cmidrule(lr){6-7}
& Before & After & Before & After & Before & After \\
& 2014-2018 & 2019-2024 & 2014-2020 & 2021-2024 & 2014-2021 & 2022-2024 \\
\midrule
Pseudocode & -0.7 & 0.5 & -1.12 & -3.1 & 3.52 & -1.44 \\
Open Code & 4.24 & 5.76 & 4.47 & 4.42 & 4.92 & 0 \\
Open Datasets & 3.48 & 1.26 & 4.66 & 1.11 & -0.38 & -0.01 \\
Dataset Splits & 2.19 & 0.41 & 5.09 & -3.15 & 0.01 & -3.27 \\
Hardware Specification & 0.51 & 7.58 & 1.05 & 1.92 & 1.62 & 2.17 \\
Software Dependencies & -0.38 & 1.1 & -0.78 & -0.86 & -0.07 & 0.68 \\
Experiment Setup & 1.19 & 1.26 & 3.17 & -1.36 & 0.29 & 0.07  \\
\midrule
\multirow{2}{*}{Reproducibility Variable} & \multicolumn{2}{c}{ICML} & \multicolumn{2}{c}{IJCAI} & \multicolumn{2}{c}{NeurIPS} \\
\cmidrule(lr){2-3} \cmidrule(lr){4-5} \cmidrule(lr){6-7}
& Before & After & Before & After & Before & After \\
& 2015-2022 & 2023-2024 & 2014-2020 & 2021-2024 & 2014-2018 & 2019-2024 \\

\midrule
Pseudocode & -0.31 & -4.44 & -2.13 & -0.35 & -0.94 & 0.59 \\
Open Code  & 5.67 & 1.42 & 4.8 & 5.2 & 5.04 & 4.91 \\
Open Datasets & 1.35 & 1.62 & 3.61 & 0.53 & 2.77 & 1.6 \\
Dataset Splits & -0.15 & -0.61 & 3.91 & -2.24 & 0.28 & 2.1 \\
Hardware Specification & 1.56 & 7.35 & -0.34 & 1.23 & 0.81 & 11.74 \\
Software Dependencies & 0.37 & -0.25 & -1.3 & 0.73 & 0 & 1.64 \\
Experiment Setup & -0.02 & -0.78 & 1.98 & -0.69 & -0.8 & 3.06 \\
\bottomrule
\end{tabular}
\caption{The slope of the trend line for each reproducibility variables before and after reproducibility checklists were introduced by the conference organizers. For all papers, we used the year 2019, when NeurIPS introduced the first reproducibility checklist of the five leading AI conferences.}
\label{tab:slopes_of_reproducibility_variables}
\end{table}

Reproducibility checklists have been introduced to improve and standardize documentation practices in the field, as checklists help establish a higher standard of baseline performance~\cite{gawande_2010} yet little information on the effect of these checklists is available.
To determine whether reproducibility checklists have had a positive effect on actual or predicted reproducibility of AI research, we analysed the correlation between the introduction of reproducibility checklists and the reproducibility variables. 
Four of the five leading AI conferences have introduced reproducibility checklists (Supplementary Table 11). 
NeurIPS was first in 2019, followed by AAAI and IJCAI in 2021, and ICML in 2023.
ICLR has introduced a guideline in 2022 to include an optional reproducibility statement.
Authors may self-select into conferences based on their requirements, including the presence of reproducibility checklists, a factor we are unable to control for.
We fitted a trend line to each reproducibility variable. 
The slope quantifies the average rate of change in percentage points per year (pp/year).
We expect to see an effect of a reproducibility checklist -- if there is any -- would appear as a change in the slope of the trend line after it is introduced.
To evaluate the impact of the reproducibility checklist, we test the hypothesis that the pp/year increased after checklist introduction, comparing pre- and post-checklist slopes using a one-sided binomial test.

Our analysis shows that reproducibility checklists did not systematically improve the documentation rates of the reproducibility variables; the trends towards better documentation practices were present before formal procedural mandates were introduced.
 
\autoref{figure:trends_all_conferences} illustrates these trends graphically for the aggregate dataset of all papers across the conferences and for each conference individually, while \autoref{tab:slopes_of_reproducibility_variables} provides the slope (pp/year) values for each reproducibility variable before and after the reproducibility checklist was introduced. 

Across the seven reproducibility variables and five conferences (n = 35), 15 exhibited a higher post-checklist pp/year (Supplementary Table 3). 
A one-sided binomial test ($H_0: p = 0.5$) indicates this is consistent with chance variation ($p = 0.84$, Cohen's $h = -0.16$), providing no statistical evidence that the adoption of the reproducibility checklist increased the rate of improvement in documentation.
The exception is \emph{Hardware specification} (purple line in \autoref{figure:trends_all_conferences}), which experiences an increase in all conferences after the introduction of reproducibility checklist. 
Although the increase is moderate for most conferences, the steep increase in \autoref{figure:trends_all_conferences} is mainly driven by NeurIPS.

\subsection{Comparison of Academia and Industry Documentation Practices}

\begin{figure}[!t]
    \centering
    \textbf{Industry affiliation rates and documentation practices relative to academia across five AI conferences from 2014 to 2024.}\\[0.5em]
    \includegraphics[width=1\linewidth]{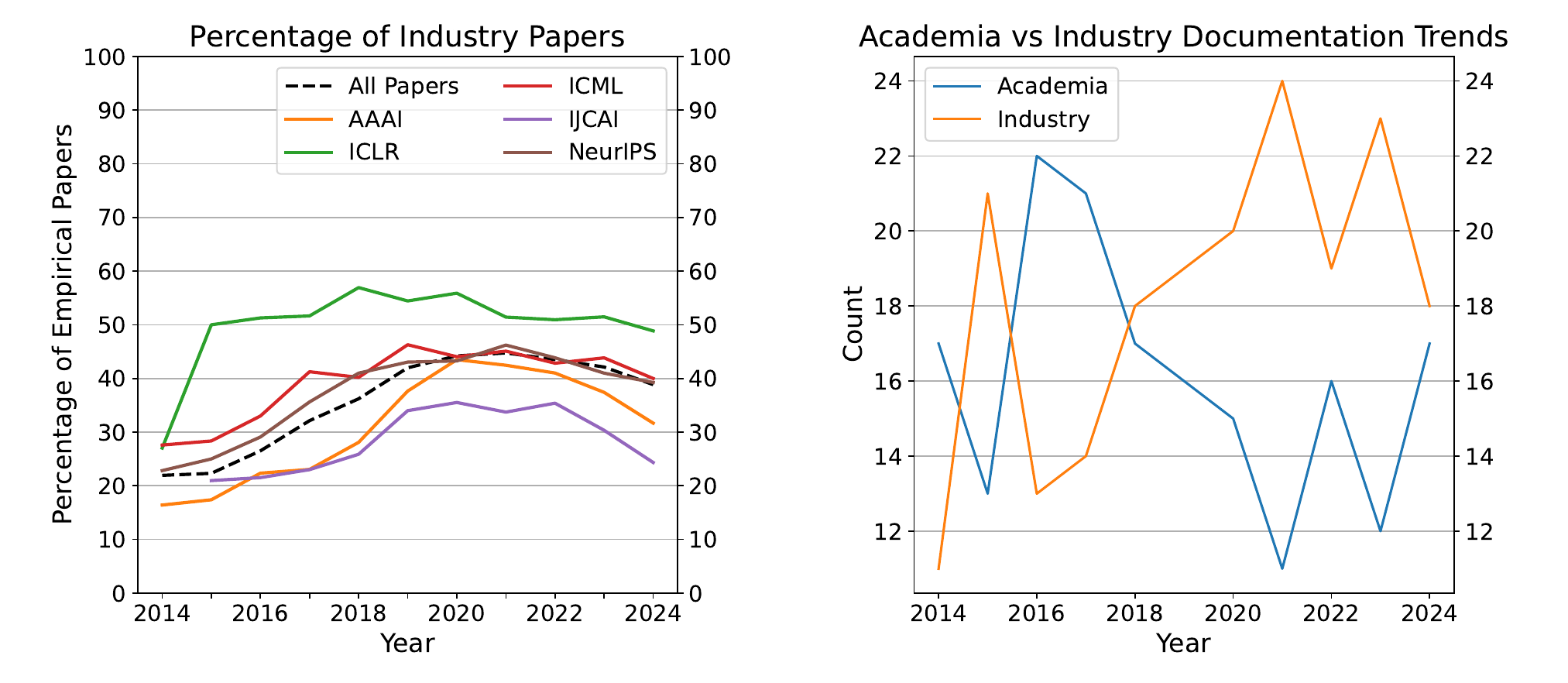}
    \caption{\textbf{Left}: The percentage of empirical papers at the five AI conferences that have at least one or more authors with industry affiliation. While the overall number of industry-affiliated authors has increased from 2014 to 2024, there has been a notable decline since the peak between 2018 and 2021. IJCAI did not hold a conference in 2014. \textbf{Right}: For each year, we counted — across all 35 variable-conference combinations (seven reproducibility variables × five conferences) — how many combinations academic papers (blue) documented at a higher rate than industry-affiliated papers, and how many industry-affiliated papers (orange) documented at a higher rate. The pattern reveals a notable shift: academic papers consistently demonstrated superior documentation practices in early years (2014-2017), but industry-affiliated papers began documenting variables more comprehensively starting in 2018, a trend that has persisted through 2024. This reversal coincides with the decline in overall industry participation.}
    \label{figure:academia_industry}
\end{figure}

Observations suggest that papers associated with industry-affiliated authors share less code and data than papers solely by academic authors~\cite{collberg_2016, gundersen_2019}. 
Industry-affiliated authors may have incentives to withhold code or data to protect intellectual property and to thus maintain a competitive advantage~\cite{collberg_2016, pineau_2021}, or may be required to do so by their employer.
To examine whether there are systematic differences in documentation practices, we evaluated which affiliation type documented each reproducibility variable at a higher rate. 
We applied one-sided binomial tests to assess whether the proportion of academic or industry papers with a higher percentage exceeded the 0.5 expected by chance if both groups were equally likely. 
Across the seven reproducibility variables, the five conferences, and 11 years (n = 378), academic papers documented more variables 177 times, while industry-affiliated papers had a higher percentage 200 times, with one instance being a statistical tie, see Supplementary Table 4. 
This result according to the binomial is consistent with chance variation and does not provide statistical evidence that there are systematic differences between academia and industry. 

Although it may appear that these findings are in disagreement with prior research, our analysis — in which industry-affiliated papers documented more variables overall, is broadly consistent with \citet{gundersen_2019}, whose analysis found academic papers to document reproducibility variables at higher rates, and indicates a shift in the trend of open code and open dataset sharing by industry-affiliated authors beginning in 2018.
When restricting our analysis to the set of conferences common to both \citet{gundersen_2019} and our study and aggregating the results following their methodology, we observe comparable patterns: academic papers document six of the seven reproducibility variables at higher rates, with dataset splits, a variable not examined by \citet{gundersen_2019}, showing a higher rate of documentation in industry papers, see Supplementary Table 5. 
Extending the analysis to the entire decade revealed a notable shift beginning in 2018, when industry-affiliated papers began to document reproducibility variables at higher rates than academic papers and continued to do so in subsequent years (\autoref{figure:academia_industry} right).
This improvement in documentation quality among industry-affiliated papers coincides with a decline in their overall publication share (\autoref{figure:academia_industry} left), suggesting that while fewer industry-affiliated papers are now being published, those that are exhibit more comprehensive documentation of reproducibility variables, a reversal of the trend observed earlier in the decade.
\citet{gundersen_2019}, based on analyses from AAAI and IJCAI conferences in 2014 and 2016, and \citet{collberg_2016}, published in 2016, both reported that industry-affiliated papers shared code less frequently than academic ones, a pattern consistent with our results, but began to change in 2018, when industry-affiliated papers started to exhibit more comprehensive documentation practices.

A possible explanation for the relative improvement in documentation quality among industry-affiliated papers is that the number of industry papers has declined since the 2018–2021 peak (\autoref{figure:academia_industry} left). 
What remains as industry submissions may be essentially academic research conducted by researchers with academic backgrounds following similar standards.
Consequently, when industry-affiliated researchers do choose to publish at conferences, they may do so with the intention of including code and data; the materials may no longer offer a competitive edge, or to align with strategic or reputational goals.

\subsection{The Prevalence of Theoretical Research}

The trend in the last decade reveals a diminishing proportion of purely theoretical work, driven by the rapid expansion of empirical AI research rather than a contraction of theoretical work in absolute numbers (Supplementary Figure 1).
Across the five leading AI conferences, the proportion of theoretical papers decreased from 10\% in 2014 to 6\% in 2024. 
This decline has been steady after a peak of 15\% in 2015.
As the field has shifted toward empirical research, open science practices have become increasingly important. 
Unlike theoretical work, where the logical and mathematical rigour of proofs is self-contained within papers, empirical research depends on access to external artifacts and explicit methodological descriptions for independent verification. 
With empirical papers now constituting 94\% of the published works at the five leading AI conferences in 2024, open science practices, such as comprehensive documentation, open source code, and open data, are now even more important to ensure the rigour, reliability, and trustworthiness of AI research.

An analysis of NeurIPS 2019 found that 9\% of the submissions were designated by the authors as mainly theoretical~\cite{pineau_2021}. 
Our LLM-based methodology, which analyses published papers rather than author-reported metadata, identified 10.3\% of accepted papers as theoretical. 
The close agreement between these estimates, despite their derivation from different data sources and methodologies, provides evidence for the validity of our approach.

\section{Discussion}

We analysed 11 years of AI research from five top-tier AI conferences to assess how documentation trends have changed and influenced the reproducibility of empirical AI studies. 
Across 56\,800 papers, documentation improved substantially over time: the share of empirical papers that provided documentation for five or more of the reproducibility variables increased from 8\% in 2014 to 43\% in 2024.
This improvement included increased sharing of source code and datasets, which raised the estimated reproducibility rate of papers from 28\% to 64\%, under the assumptions described in \autoref{sec:improved-documentation-and-open-science-practices}.
Statistical analysis revealed no evidence that the introduction of reproducibility checklists accelerated these changes. 
The upward trend in documentation started several years before the introduction of reproducibility checklists, indicating that broader shifts toward open science and improved documentation practices either preceded formal reproducibility requirements at these conferences or that formal requirements emerged as a response to the same underlying trends driving improved documentation practices.

Given this progress, a critical question emerges: can this trajectory be sustained?
The slowing growth in our estimated reproducibility rate in recent years suggests the community may be approaching practical limits. One limit is the use of closed datasets, which accounted for 9\% of empirical papers in 2024. 
Achieving 100\% open data is neither realistic nor desirable, as legitimate privacy and confidentiality concerns renders in some cases openly sharing data unrealistic.

A more tractable frontier for improvement lies with the 27\% of papers that use open datasets but do not share code. \citet{gundersen_2025} finds one third of these papers to be reproducible; closing this gap represents the most direct path to increasing the estimated reproducibility rate as the mentioned concerns of sharing of data rarely applies to code. 
Intellectual property is cited as a concern for researchers working in industry~\cite{collberg_2016, pineau_2021}.
As AI research has become increasingly profitable, industry may have become more reluctant to publish work that is closest to actual applications to maintain a competitive advantage.
However, if the implementation is described in sufficient detail to enable reproduction, code sharing reveals no additional proprietary information while reducing barriers to building on existing work, though this efficiency gain may diminish authors' competitive advantage in pursuing follow-up research.

Withholding code reduces transparency and hampers reproducibility, as code often contains crucial implementation details that may not be documented. 
Although there may be good reasons for withholding both code and data, the default should be an expectation to share.
Those who cannot share should be asked to state why.

Overcoming these challenges to push beyond the current plateau will require a more fundamental evolution in how we author and publish AI research. 
The principles behind initiatives such as the "Geoscience Paper of the Future", articulated by \citet{gil_2016}, would strengthen AI reproducibility by ensuring that data and software are reusable, properly licensed, and persistently identifiable, while computational workflows are explicitly documented to capture provenance and support transparent verification and reuse.
This is critical not just for human researchers, but for the next generation of AI research. As argued by \citet{gil_2022}, published papers are often too unstructured for automated analysis and reproduction. 
Tools that analyse papers, code, and data -- such as Reproscreener~\cite{bhaskar_2024}, which automatically assesses computational reproducibility -- and Reproducibility Copilots -- such as OpenPub~\cite{bibal_2025}, which uses AI to facilitate replication by generating Jupyter Notebooks with code and actionable recommendations -- demonstrate the use of AI to work around the challenges in extracting reproducibility information from traditional paper formats. 
These tools must work around incomplete or unstructured documentation and would benefit from an approach to publication that prioritizes documentation and open artifacts. 
By making publications machine-readable, we facilitate the development of AI systems that can independently reproduce, extend, and ultimately generate new scientific discoveries on their own. 
Researchers have already developed tools to help scientists generate novel hypotheses and research proposals, such as Google AI co-scientist~\cite{gottweis_2025}; although not without limitations~\cite{cheetham2024artificial}, it has already shown to be valuable~\cite{guan2025assisted, he2025chimeric}.

Our study has several limitations.
First, our LLM-based analysis relied exclusively on the plain text extracted from papers, excluding information potentially contained in figures, tables, supplementary materials, or external repositories not mentioned in the text. 
Second, our assessment of reproducibility variables is binary; it does not capture the quality or completeness of the shared artifacts. For example, a paper sharing partial code is treated the same as one with a complete, well-documented repository. 
Third, LLMs can exhibit generalisation bias~\cite{peters_2025}, and while our prompts were designed to mitigate this, a subtle influence on the model cannot be ruled out.
Finally, our estimation of reproducibility is based on the availability of code and data and is highly uncertain. 
The numbers on which we base the analysis come from a small study that evaluated whether the authors were able to draw the same conclusions from the same data.
The study did not evaluate whether the claims can be generalised to new data.
Still, these are the best source data that we have found to base a field at scale estimation of reproducibility from. 

Future work should focus on addressing these limitations, and on expanding the scope of inquiry. 
Methodologically, future tools could move beyond binary classifications and plain text to assess documentation quality, parse code repositories, and interpret figures. 
Additional descriptive analyses, breakdowns by institution type, subfield, and geographic region, and a before/after analysis centered on the public release of AI coding assistants would complement the longitudinal trends we report. 
Expanding the analysis to include more conferences and journals would enable a more rigorous examination of researcher self-selection into venues with or without reproducibility checklists, and could support an empirical estimate of the probability of reproducibility when all seven variables are documented.

Furthermore, comparing trends in AI to other scientific disciplines would provide valuable context, revealing whether this embrace of open science is unique to AI or part of a larger movement.


\section{Methods}

\subsection{Reproducibility Variable Selection}
\label{sec:reproducibility-variable-selection}

The 20 variables identified by \citet{gundersen_2018} provide the basis for our selection of the reproducibility variables used to assess the progress of open science in AI research. 
\citet{gundersen_2018} provided both the conceptual framework and, critically, a prompt optimization dataset of 400 manually annotated papers. 
Building on this existing dataset allowed us to begin prompt engineering without first investing substantial time and resources in manual annotation, and the breadth of the variable set covers the major dimensions of empirical AI research documentation — method, data, experiment, and metadata — making it a well-motivated starting point.
These include aspects of the method documentation (problem, objective/goal, research method, research questions, pseudocode), data documentation (training data, validation data, test data, results), experiment documentation (hypothesis, prediction, method source code, hardware specifications, software dependencies, experiment setup, experiment source code), and miscellaneous information about the research (research type, research outcome, affiliation, and contribution). 
The necessity of documenting these variables is rooted in the principle that sufficient detail must be provided for an independent team to reproduce the work. 
The variables are primarily Boolean, with one binary variable, research type (empirical/theoretical), and one ternary variable, author affiliation (academia/industry/collaboration).

A subsequent replication study by \citet{gundersen_2025} reinforced the critical importance of sharing code and data. 
This finding prompted us to refine the original data-related variables. 
Instead of checking for separate training, validation, and test data sharing, we replaced them with two new reproducibility variables, reducing the effective variable set from 20 to 19:

\begin{itemize}
    \item Open Datasets: This is true if the paper uses a well-known public dataset or shares the dataset via a URL, DOI, or formal citation.
    \item Dataset Splits: This is true if the paper specifies how data is split, whether through exact percentages, absolute counts, predefined splits, stratified methods, or cross-validation.
\end{itemize}

To select a robust subset of variables for our automated analysis, we applied three criteria. 
First, we required class balance, excluding variables where one class had too few instances for reliable LLM evaluation; the first criterion eliminated seven variables, Result Outcomes, Research Method, Research Question, Hypothesis, Prediction, Open Experiment Code, and Results, from consideration.
Supplementary Table 6 shows the class balance for each reproducibility variable.
Second, we prioritized variables deemed most important for reproducibility based on the findings of \citet{gundersen_2025}, which emphasised that providing code and data are paramount. 
Accordingly, we classified pseudocode, open datasets, dataset splits, results, open code, hardware specification, software dependencies, experiment setup, and open experiment code as important. 
We also included affiliation and research type as important variables, as their results may influence whether code and data are shared. 
Third, we required the LLM to achieve an $F_1$ score of 75\% or greater for the reproducibility variable during our prompt engineering phase on the prompt optimisation dataset; this threshold was selected to ensure reliable classification while retaining sufficient variables for a comprehensive analysis.
Supplementary Table 2 shows the $F_1$ scores for the reproducibility variables on the prompt optimisation dataset.
The three remaining excluded variables, Problem Description, Goal/Objective, and Contribution, passed the class balance criterion but did not achieve F1 scores of 75\% or greater on the prompt optimization dataset. 
These variables are not always stated explicitly using consistent terminology in papers and require a higher level of contextual understanding than we were able to achieve reliably through prompting on the LLMs we tested. 
The excluded variables are not necessarily less important in principle; rather, their reliable automated extraction was not achievable within our current methodology, and we chose not to include any variable for which we lacked confidence in the results.

We made modifications to the original dataset from \citet{gundersen_2018} to better suit the capabilities of our LLM-based evaluation. 
The open code variable was reevaluated; the original study verified if the code was still accessible at the provided link, a task current LLMs cannot perform. 
Our revised criterion only checks if the paper mentions a link to the code. 
We also reevaluated software dependencies to ensure the criterion was consistently applied, requiring the mention of both a software package and its version number. 
Justifications, with quotes from the papers, for the changes to open datasets, dataset splits, open code and software dependencies are provided in supplementary material.
For the affiliation variable, we noticed that the original dataset contained only 11 papers in the industry category (i.e., with all authors from industry).
With such a small sample size, we would not have been able to reliably assess the ability of our method to detect this type of papers. 
We thus used a binary classification instead: papers written by authors solely from academia \emph{vs} ones with at least one author affiliated with industry.

\subsection{Automated Evaluation of Reproducibility Variables}

To develop and validate the LLM prompts used for our automated evaluation method, we used the dataset created by \citet{gundersen_2018} containing 400 papers from AAAI (2014 and 2016) and IJCAI (2013 and 2016) and manually evaluated for their 20 reproducible variables, which we refer to as the prompt optimisation dataset. 
We downloaded PDF versions of the 400 papers from the AAAI and IJCAI proceedings websites and extracted just the text of the papers, excluding figures but including figure captions and the text of the tables, without the table formatting. 
Then we created a script to automate the process of submitting the paper text and the reproducibility variable evaluation prompt to the APIs for Anthropic, Google, OpenAI and open weight models for inference and save the results in a structured format.   

We undertook an iterative prompt engineering process to optimise LLM accuracy on each of the 20 reproducibility variables with the prompt optimisation dataset. 
We did not modify the weights of the model or use other techniques, such as Retrieval-Augmented Generation (RAG). 
Instead, we utilised a few-shot prompting technique. 
A few-shot prompt provides a few natural language demonstrations of the task at inference time, but no weight updates are performed~\cite{brown_2020}.
As the papers analysed in this study were almost certainly used to train the LLMs we evaluated, the degree to which responses reflect in-context reasoning versus recall remains an open question; the method nonetheless achieved F1 scores exceeding 90.0\% for six of nine variables on the Evaluation Dataset (Supplementary Table 9).
For each variable, we drafted a few-shot prompt and used it to evaluate a subset of the 400 papers. 
For each reproducibility variable we asked the LLM to return two values: (1) the Boolean, binary, or ternary result of the engineered prompt and (2) a string result that includes a quote of the text from the paper that supports the first result, otherwise explain why the information in the paper is insufficient.
The classifications obtained from the LLM were then compared against the ground truth of the prompt optimisation dataset using a confusion matrix. 
We closely examined the false positives and false negatives, reviewing the quotes from the LLMs string output with the source text of the papers to identify patterns of error. 
This analysis informed subsequent revisions of the few-shot prompt and the cycle was repeated until the performance of the LLM stabilised at a high level of accuracy and $F_1$ score.
Examples demonstrating the model's handling of negated statements, such as explicit declarations that a dataset is not publicly available, are provided in Supplementary Section S6.

The final prompt submitted to the model for each paper consisted of a single API query containing a brief task description followed by the full plain text of the paper and the specific few-shot prompts for each of the 20 variables. 
The prompts for each variable asked the LLM to determine if the paper met the criterion, provided a list of positive and negative examples to guide its decision, requested a direct quote from the paper to support its answer, and concluded with a final yes/no question. 
The model was instructed to return its output in a structured JSON format, containing its binary answer and the supporting text from the paper.
The full prompt, excluding the text of the individual papers due to size limitations, is available in supplementary material.

\subsection{Model Selection and Prompt Engineering} 

We evaluated 17 models from Anthropic, Google, OpenAI, Alibaba Cloud, and Microsoft, of which three were open weight models (Supplementary Table 8).
We selected the Google Gemini 2.5 Flash model for the final large-scale evaluation based on four key criteria:

\begin{description}
\item [LLM API Features:] The task required reliable structured output. 
The Google, OpenAI, and Ollama (used with the open weight models) APIs support generating JSON output based on a predefined schema, ensuring consistency. 
The Anthropic API lacked this native capability and was therefore eliminated. 
\item [Time:] Due to limited access to GPU resources, we excluded the three open weights models, since we could not process 58\,600 papers within a reasonable timeframe. 
\item [Cost:] Reserving funds for reasoning and output tokens, which are more expensive and difficult to predict the length of, we selected models that would fit our budget for input tokens for all 56\,800 papers, based on an estimated 13\,000 input tokens per paper (Estimated costs are in Supplementary Table 8).
\item [Accuracy:] In preliminary tests, Gemini 2.5 Flash demonstrated superior accuracy on our task compared to other models within our budget constraints, a result corroborated by the public LLM leaderboard LMArena (\href{https://lmarena.ai/}{https://lmarena.ai/})~\cite{chiang_2024}.
\end{description}

The LLM-generated results for each of the 400 papers were systematically evaluated against the ground truth from the prompt optimisation dataset over five runs. 
We calculated the mean accuracy and mean $F_1$ score for each reproducibility variable to quantify model performance and selected affiliation, research type, pseudocode, open datasets, dataset splits, open code, hardware specification, software dependencies, and experiment setup as the final variables for our large-scale analysis. The results for each run can be found in Supplementary Table 7.

\begin{figure}[!t]
    \centering
    \textbf{LLM classification consistency across five runs for each reproducibility variable.}\\[0.5em]
    \includegraphics[width=1\linewidth]{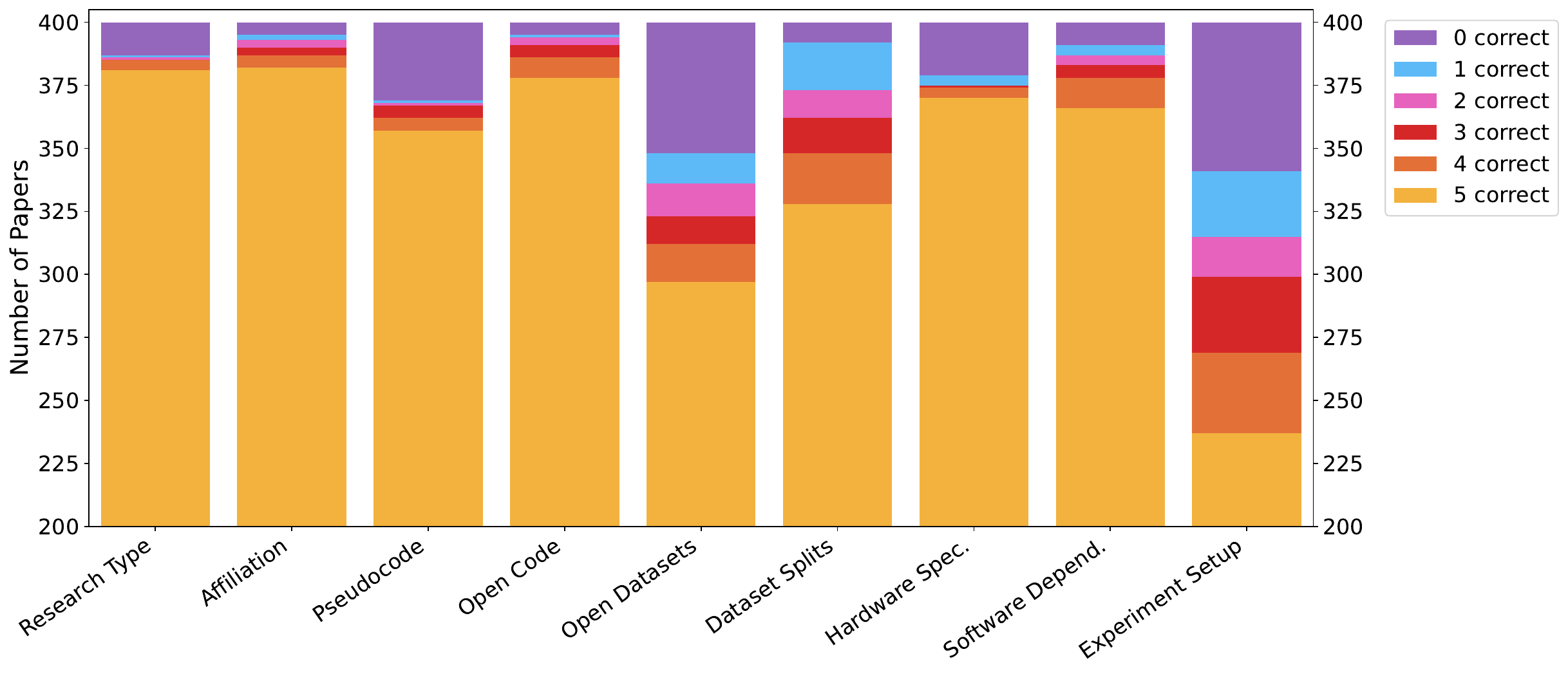}
    \caption{The consistency of classification from the LLM over five runs for each of the reproducibility variables. Light orange represents the number of papers the LLM consistently correctly classified reproducibility variable and the purple represents the number of papers the LLM consistently incorrectly classified reproducibility variable. The dark orange, red, pink and blue, represents the number of papers the LLM did not consistently make the same classification over multiple runs.}
    \label{figure:accuracy_consistency}
\end{figure}

Google Gemini 2.5 Flash demonstrated high performance and stability across five evaluation runs. 
For affiliation and pseudocode, mean $F_1$ scores greater than 90\% was achieved.
For research type and hardware specification, open code, open datasets, and dataset splits, we found $F_1$ scores greater than 80\%.
Software dependencies and experiment setup were also predicted reliably, with $F_1$ scores of 78.3\% and 78.8\%, respectively.
The variance across runs differed substantially for the different variables; the accuracy range varied from 0.5\% (hardware specification) to 4.8\% (dataset splits), and the $F_1$ score range varied from 1.0\% (hardware specification) to 11.9\% (software dependencies) -- see Supplementary Table 7 for all means and ranges.
Furthermore, the consistency of classifications for individual papers was high. 
For research type, affiliation, pseudocode, open code, hardware specification, and software dependencies, the evaluation was identical across all five runs over 95\% of the time. 
Open dataset, dataset splits, and experiment setup also showed high consistency, with consistent evaluations 87\%, 84\%, and 74\% of the time, respectively, see \autoref{figure:accuracy_consistency}.

\subsection{Large-Scale Conference Paper Analysis}

For our large-scale analysis, we selected five top-tier AI conferences: AAAI, ICLR, ICML, IJCAI, and NeurIPS. 
We collected all published papers over an 11-year period to analyse trends over time. 
After excluding papers from affiliated workshops and papers from joint proceedings, which often have different reviewing criteria, our final dataset comprised 56\,800 main conference papers. 
We did not include journals in our analysis. Although these journals, such as JAIR and JMLR, play major roles in AI research, we excluded these due to time constraints.
This set also included 300 papers from the original analysis by \citet{gundersen_2018} (AAAI 2014, AAAI 2016, and IJCAI 2016). 
A detailed description of the paper selection and collection process is available in Supplementary Section S3.

The workflow for processing the 56\,800 papers mirrored the one used for the initial 400-paper evaluation. 
Each PDF was converted to text, and the same engineered prompt was used with Google Gemini 2.5 Flash
for all of the reproducibility variables.
Only the nine selected reproducibility variables from \autoref{sec:reproducibility-variables} were analysed. 
The results from the 56\,800 paper LLM analysis containing the binary answer and the supporting text from the paper are provided as supplementary material.

\subsection{Does the Method Generalise to the Population?}

To validate how well our approach generalised from the prompt optimisation dataset to the full set of papers, we sampled 160 papers uniformly at random from full set of 56\,800 papers and labelled them manually to create an evaluation dataset. 
The results are presented in Supplementary Table 9.
The year and conference distributions of the evaluation dataset closely mirror those of the full corpus (Supplementary Figure 2 and Supplementary Figure 3), supporting the generalisability of the results reported below.
Surprisingly, the $F_1$ scores show a substantial improvement for all classes, achieving greater than $90.0\%$ with the exception of three variables; data set splits ($81.48\%$), software dependencies ($81.25\%$) and experiment setup ($86.21\%$). This increase in performance can be attributed to various factors. First, the prompt optimisation dataset is a biased subset of the population, representing only AAAI and IJCAI in 2013, 2014 and 2016. Secondly, the LLM was not actually fitted to the training data, but rather the prompts were optimised, 
which prevents possible overfitting through backpropagation. Only for one variable a decrease in performance was measured between the prompt optimisation  and evaluation datasets: for dataset splits performance dropped from $81.48\%$ to $79.56\%$, a decrease of 1.92 percentage points. 
We found that the majority of the papers in the evaluation set that had mislabelled data splits were caused by inflexibility in the prompt; in 60.0\% of the errors, the procedure produced a false negative due to the authors choosing not to use a validation set and therefore not providing the full data split containing training, validation and test sets. 
Overall, the results show us that our method generalises well to the full dataset of papers.


\backmatter

\bmhead{Supplementary information}

The LLM prompt, code, results, and analysis can be found online: \href{https://doi.org/10.5281/zenodo.20785801}{https://doi.org/10.5281/zenodo.20785801}~\cite{coakley_software_2026}.

\bmhead{Acknowledgements}

Cloud computing services provided by CloudBank: National Science Foundation [\# 1925001]. 

\section*{Declarations}

\subsection{Author contributions}

K.L.C. conceived and designed the experiments, performed the experiments, analysed the data, contributed materials/analysis tools, and wrote the paper.
T.S. analysed the data, contributed materials/analysis tools, and wrote the paper.
H.H. wrote the paper.
O.E.G. conceived and designed the experiments, analysed the data, contributed materials/analysis tools, and wrote the paper.

\subsection{Competing interest}

The authors declare no competing interests.

\subsection{Data availability}

The datasets generated during the current study are available in the Zenodo repository, \href{https://doi.org/10.5281/zenodo.20785801}{https://doi.org/10.5281/zenodo.20785801}~\cite{coakley_software_2026}.
We do not include the PDFs for of all 56\,800 papers analysed for this study due to copyright concerns.
The Zenodo repository includes the code to download and preprocess all 56\,800 papers.

\subsection{Code availability}

The code used for the experiment, include preprocessing and analysis, is available in the Zenodo repository, \href{https://doi.org/10.5281/zenodo.20785801}{https://doi.org/10.5281/zenodo.20785801}~\cite{coakley_software_2026}.

\subsection{Funding declaration}

K.L.C discloses support for the research of this work from National Science Foundation [\# 2226453]. O.E.G. discloses support for the research of this work from RICO (Robust Intelligent Control, Research Council of Norway) [\# 329730]. T.S. and H.H. discloses support for the research of this work from through an Alexander von Humboldt Professorship held by Holger Hoos from the Alexander von Humboldt Foundation.

\clearpage
\bibliography{sn-bibliography}


\clearpage

\renewcommand{\figurename}{Supplementary Figure} 
\renewcommand{\figureautorefname}{Supplementary Figure} 
\renewcommand{\tablename}{Supplementary Table}        
\renewcommand{\tableautorefname}{Supplementary Table} 
\renewcommand{\thesection}{S\arabic{section}}

\setcounter{figure}{0} 
\setcounter{table}{0}  

\section{Supplementary Tables}

\begin{table}[htbp]
\centering
\begin{tabular}{lccccc}
\toprule
\textbf{Reproducibility Variable} & \textbf{AAAI} & \textbf{ICML} & \textbf{ICLR} & \textbf{IJCAI} & \textbf{NeurIPS} \\
 & \textbf{2021} & \textbf{2023} & \textbf{2022} & \textbf{2021} & \textbf{2019} \\
\midrule
Pseudocode           & \checkmark & -- & -- & \checkmark & -- \\
Open Code            & \checkmark & \checkmark & \checkmark & \checkmark & \checkmark \\
Open Datasets         & \checkmark & \checkmark & \checkmark & \checkmark & \checkmark \\
Dataset Splits       & -- & \checkmark & -- & \checkmark & \checkmark \\
Hardware Specification             & \checkmark & \checkmark & -- & \checkmark & \checkmark \\
Software Dependencies        & \checkmark & -- & -- & \checkmark & \checkmark \\
Experiment Setup   & \checkmark & \checkmark & -- & \checkmark & \checkmark \\
\bottomrule
\end{tabular}
\caption{Reproducibility variables specified in conference reproducibility checklists. The year each checklist was introduced is shown beneath the conference name. Reproducibility variables are described in \textbf{Section 1.1}. Checkmarks indicate which conferences require each variable in their submission guidelines.}
\label{tab:conference_checklists}
\end{table}

\begin{table}[h!]
\centering
\begin{tabular}{ccc|ccc}
\toprule
\textbf{Imbalanced Classes} & \textbf{Accuracy} & \textbf{$F_1$Score} & \textbf{Balanced Classes} & \textbf{Accuracy} & \textbf{$F_1$ Score} \\

\midrule
Open Experiment Code & 94.6\% & 55.5\% & \textbf{Affiliation} & \textbf{97.4\%} & \textbf{92.2\%} \\ 
Results & 96.7\% & 30.8\% & \textbf{Research Type} & \textbf{96.2\%} & \textbf{88.7\%} \\
Result Outcome & 91.9\% & 95.8\% & \textbf{Pseudocode} & \textbf{91.2\%} & \textbf{90.4\%} \\
Research Method & 96.7\% & 5.7\% & \textbf{Open Code} & \textbf{97.2\%} & \textbf{81.8\%} \\
Research Question & 93.5\% & 52.5\% & \textbf{Open Datasets} & \textbf{80.8\%} & \textbf{81.2\%} \\ 
Hypothesis & 94.1\% & 54.3\% & \textbf{Dataset Splits} & \textbf{90.2\% }& \textbf{83.4\%} \\
Prediction & 91.3\% & 14.8\% & \textbf{Hardware Specification} & \textbf{93.7\%} & \textbf{86.8\%} \\
& & & \textbf{Software Dependencies} & \textbf{95.3\%} & \textbf{78.3\%} \\
& & & \textbf{Experiment Setup} & \textbf{73.1\%} & \textbf{78.8\%} \\
& & & Problem Description & 58.9\% & 65.7\% \\
& & & Goal Objective & 81.9\% & 65.7\% \\\
& & & Contribution & 68.7\% & 74.9\% \\
\bottomrule
\end{tabular}
\caption{Performance of the LLM-based automated method on reproducibility variables from \citet{gundersen_2018}. Mean accuracy and $F_1$ scores are shown across five evaluation runs. Variables are grouped by class balance, with "Imbalanced Classes" containing variables where one class has fewer than 25 instances (see \autoref{tab:reproducibility_variables_label_distribution} for distributions). The nine bolded variables were selected for the large-scale analysis based on three criteria: balanced class representation, importance for reproducibility, and $F_1$ scores exceeding 75\%.}
\label{tab:analysis_of_reproducibility_variables}
\end{table}

\begin{table}[htbp]
\centering
\begin{tabular}{lccccc}
\toprule
\textbf{Reproducibility Variable} & \textbf{AAAI} & \textbf{ICML} & \textbf{ICLR} & \textbf{IJCAI} & \textbf{NeurIPS} \\
\midrule
Pseudocode & --	& -- & -- & \checkmark & \checkmark \\
Open Code & -- & -- & --	& \checkmark & -- \\
Open Datasets & -- & \checkmark & \checkmark & -- & -- \\
Dataset Splits & -- & -- & -- & -- & \checkmark \\
Hardware Specification & \checkmark & \checkmark & \checkmark & \checkmark & \checkmark \\
Software Dependencies & -- & \checkmark & -- & \checkmark & \checkmark \\
Experiment Setup & -- & -- & -- & -- & \checkmark \\
\bottomrule
\end{tabular}
\caption{Summary of directional changes in documentation trends before and after the introduction of reproducibility checklists. For each reproducibility variable and conference, a checkmark indicates that the post-checklist slope (pp/year) exceeds the pre-checklist slope; ``--'' indicates no increase. Of the 35 variable-conference pairs, 15 showed an increase (one-sided binomial test, $p = 0.84$, Cohen's $h = -0.16$), providing no statistical evidence that checklist adoption accelerated documentation improvement. Corresponding slope estimates are reported in \textbf{Table 1}.}
\label{tab:conference_checklists_binomial_tests}
\end{table}

\begin{table}[t]
\centering
\begin{tabular}{lccccccccccc}
\toprule
\textbf{Reprod. Variable} & \textbf{2014} & \textbf{2015} & \textbf{2016} & \textbf{2017} & \textbf{2018} & \textbf{2019} & \textbf{2020} & \textbf{2021} & \textbf{2022} & \textbf{2023} & \textbf{2024} \\
\midrule
\multicolumn{12}{l}{AAAI} \\
\midrule
Pseudocode & A & I & A & A & A & A & A & A & A & A & A \\
Open Code & A & A & A & A & A & A & A & A & A & A & A \\
Open Datasets &A & I & A & I & I & I & I & I & I & I & I \\
Dataset Splits & I & A & I & I & I & I & I & I & I & I & I \\
Hardware Specification & A & I & A & I & I & A & A & A & A & I & A \\
Software Dependencies & A & A & A & A & A & A & A & A & A & A & A \\
Experiment Setup &I & I & A & A & I & I & I & I & A & I & I \\
\midrule
\multicolumn{12}{l}{ICLR} \\
\midrule
Pseudocode & A & Tie & A & A & I & A & A & I & I & A & A \\
Open Code & A & I & A & A & A & A & A & I & A & A & A \\
Open Datasets &A & A & A & A & A & I & I & I & I & I & I \\
Dataset Splits & I & I & I & A & I & I & I & I & I & I & I \\
Hardware Specification & A & I & A & A & I & A & I & I & I & I & I \\
Software Dependencies & A & I & I & A & A & A & I & I & A & A & I \\
Experiment Setup &I & I & A & A & I & I & I & I & I & I & I \\
\midrule
\multicolumn{12}{l}{ICML} \\
\midrule
Pseudocode & I & I & A & A & A & I & I & A & A & A & A \\
Open Code & A & A & A & A & A & A & I & A & A & A & A \\
Open Datasets &A & I & I & I & I & I & I & I & I & I & I \\
Dataset Splits & I & I & I & I & I & I & I & I & I & I & I \\
Hardware Specification & I & I & A & I & I & I & I & I & I & I & I \\
Software Dependencies & A & I & A & A & I & I & I & I & A & I & I \\
Experiment Setup &I & I & A & I & A & I & I & I & I & I & I \\
\midrule
\multicolumn{12}{l}{IJCAI} \\
\midrule
Pseudocode & -- & A & I & I & A & A & A & I & A & I & A \\
Open Code & -- & A & I & A & A & I & I & A & A & A & A \\
Open Datasets &-- & A & A & I & A & A & A & I & I & I & A \\
Dataset Splits & -- & I & I & I & I & I & I & I & I & I & I \\
Hardware Specification & -- & A & A & A & A & A & A & A & A & I & I \\
Software Dependencies & -- & A & A & A & A & A & A & A & A & A & A \\
Experiment Setup &-- & I & I & A & I & A & A & I & I & I & A \\
\midrule
\multicolumn{12}{l}{NeurIPS} \\
\midrule
Pseudocode & I & A & A & A & A & A & A & A & A & A & A \\
Open Code & A & A & I & A & A & I & A & I & A & A & A \\
Open Datasets &I & I & I & I & I & I & I & I & I & I & I \\
Dataset Splits & A & I & I & I & I & I & I & I & I & I & I \\
Hardware Specification & A & I & I & I & I & I & I & I & I & I & A \\
Software Dependencies & A & A & A & A & A & I & A & I & I & I & A \\
Experiment Setup &I & I & A & I & I & A & A & A & I & I & I \\
\bottomrule
\end{tabular}
\caption{Comparison of academic and industry-affiliated authors’ documentation of reproducibility variables across AI conferences (2014–2024). Each cell indicates whether academic (A) or industry (I) papers had a higher percentage of documentation for a given reproducibility variable and year; “Tie” indicates no difference. One-sided binomial tests ($H_0: p = 0.5$, ties excluded, $n = 377$) find no statistically significant systematic difference favouring academia ($p = 0.89$, Cohen's $h = 0.05$) or industry ($p = 0.13$, Cohen's $h = 0.05$)}
\label{tab:academia_vs_industry_binomial_tests}
\end{table}

\begin{table}[htbp]
\centering
\begin{tabular}{lrr}
\toprule
\textbf{Reproducibility Variable} & \textbf{Academia} & \textbf{Industry} \\
\midrule
Pseudocode & \textbf{46.9\%} & 44.7\%  \\
Open Code & \textbf{10.9\%} & 9.9\% \\
Open Datasets & \textbf{67.9\%} & 63.3\% \\
Dataset Splits & 28.5\% & \textbf{34.2\%} \\
Hardware Specification & \textbf{26.1\%} & 23.0\% \\
Software Dependencies & \textbf{10.7\%} & 5.8\% \\
Experiment Setup & \textbf{76.9\%} & 75.7\% \\
\bottomrule
\end{tabular}
\caption{Comparison of documentation rates for reproducibility variables between academic and industry-affiliated papers at AAAI 2014, AAAI 2016, and IJCAI 2016, following the methodology of \citet{gundersen_2019}. Percentages indicate the proportion of papers documenting each variable within each affiliation group. Bold values indicate the affiliation group with the higher documentation rate for each reproducibility variable.}
\label{tab:academia_vs_industry_gundersen_2019}
\end{table}

\begin{table}[h!]
\begin{tabular}{lcccccc}
\toprule
& \shortstack{Research\\Type} & \shortstack{Result\\Outcome} & Affiliation & \shortstack{Problem\\Description} & \shortstack{Goal/\\Objective} \\
\midrule
True & 75 (Theoretical) & 377 (Positive) & 69 (Industry/Collab) & 186 & 81 & \\
False & 325 (Empirical) & \textbf{23} (Negative) & 331 (Academia) & 214 & 319 & \\
Total & 400 & 400 & 400 & 400 & 400 & \\
\midrule 
& \shortstack{Research\\Method} & \shortstack{Research\\Question} & Hypothesis & Prediction & Contribution & \\
\midrule
True & \textbf{5} & \textbf{20} & \textbf{17} & \textbf{4} & 187 & \\
False & 395 & 380 & 383 & 396 & 213 & \\
Total & 400 & 400 & 400 & 400 & 400 & \\
\midrule 
& Pseudocode & \shortstack{Open\\Source Code} & \shortstack{Open\\Experiment Code} & \shortstack{Open\\Datasets} & \shortstack{Dataset\\Splits} & \\
\midrule
True & 177 & 29 & \textbf{18} & 171 & 126 & \\
False & 148 & 296 & 307 & 154 & 199 & \\
Total & 325 & 325 & 325 & 325 & 325 & \\
\midrule 
& Results & \shortstack{Hardware\\Specification} & \shortstack{Software\\Dependencies} & \shortstack{Experiment\\Setup} & & \\
\midrule
True & \textbf{12} & 89 & 46 & 223 & & \\
False & 313 & 236 & 279 & 102 & & \\
Total & 325 & 325 & 325 & 325 & & \\
\bottomrule
\end{tabular}
\caption{Distribution of results for the reproducibility variables across 400 manually evaluated \textit{State of the Art: Reproducibility in Artificial Intelligence} papers. Reproducibility variables with fewer than 25 instances (in Bold) of either class were excluded from the large-scale conference paper analyses due to class imbalance. The final two rows include only 325 manual evaluations, as they apply only to empirical papers.}
\label{tab:reproducibility_variables_label_distribution}
\end{table}

\begin{table}[h!]
\centering
\begin{tabular}{l*{5}{rr}}
\toprule
\multirow{2}{*}{Run} & \multicolumn{2}{c}{Research Type} & \multicolumn{2}{c}{Affiliation} & \multicolumn{2}{c}{Pseudocode} & \multicolumn{2}{c}{Open Code} & \multicolumn{2}{c}{Open Datasets} \\
\cmidrule(lr){2-3} \cmidrule(lr){4-5} \cmidrule(lr){6-7} \cmidrule(lr){8-9} \cmidrule(lr){10-11}
& Acc & $F_1$ & Acc & $F_1$ & Acc & $F_1$ & Acc & $F_1$ & Acc & $F_1$ \\
\midrule
1 & 96.3\% & 88.9\% & 97.8\% & 93.3\% & 91.3\% & 90.5\% & 97.5\% & 83.9\% & 80.5\% & 81.0\% \\
2 & 96.3\% & 88.9\% & 97.5\% & 92.7\% & 91.0\% & 90.2\% & 97.8\% & 84.8\% & 80.3\% & 80.7\% \\
3 & 96.0\% & 88.1\% & 97.3\% & 91.9\% & 90.5\% & 89.7\% & 97.3\% & 82.5\% & 83.5\% & 83.5\% \\
4 & 95.8\% & 87.2\% & 97.0\% & 91.4\% & 91.3\% & 90.5\% & 97.0\% & 80.7\% & 80.0\% & 80.8\% \\
5 & 96.8\% & 90.5\% & 97.3\% & 91.9\% & 91.8\% & 91.0\% & 96.5\% & 77.4\% & 79.8\% & 80.2\% \\
\midrule
Range & 1.0\% & 3.3\% & 0.8\% & 1.9\% & 1.3\% & 1.2\% & 1.3\% & 7.3\% & 3.8\% & 3.3\% \\
Median & 96.3\% & 88.9\% & 97.3\% & 91.9\% & 91.3\% & 90.5\% & 97.3\% & 82.5\% & 80.3\% & 80.8\% \\
Mean & 96.2\% & 88.7\% & 97.4\% & 92.2\% & 91.2\% & 90.4\% & 97.2\% & 81.8\% & 80.8\% & 81.2\% \\
\bottomrule
\end{tabular}

\begin{tabular}{l*{4}{rr}} 
\multirow{2}{*}{Run} & \multicolumn{2}{c}{Dataset Splits} & \multicolumn{2}{c}{Hardware Specification} & \multicolumn{2}{c}{Software Dependencies} & \multicolumn{2}{c}{Experiment Setup} \\
\cmidrule(lr){2-3} \cmidrule(lr){4-5} \cmidrule(lr){6-7} \cmidrule(lr){8-9}
& Acc & $F_1$ & Acc & $F_1$ & Acc & $F_1$ & Acc & $F_1$ \\
\midrule
1 & 91.5\% & 86.0\% & 93.8\% & 86.9\% & 95.3\% & 77.7\% & 73.0\% & 78.7\% \\
2 & 92.3\% & 86.8\% & 93.5\% & 86.7\% & 94.5\% & 74.4\% & 73.8\% & 79.6\% \\
3 & 89.5\% & 82.5\% & 93.5\% & 86.5\% & 95.0\% & 78.3\% & 72.8\% & 78.6\% \\
4 & 87.5\% & 78.5\% & 94.0\% & 87.5\% & 94.5\% & 75.0\% & 72.0\% & 78.0\% \\
5 & 90.0\% & 83.1\% & 93.5\% & 86.6\% & 97.0\% & 86.4\% & 73.8\% & 79.1\% \\
\midrule
Range & 4.8\% & 8.4\% & 0.5\% & 1.0\% & 2.5\% & 11.9\% & 1.8\% & 1.6\% \\
Median & 90.0\% & 83.1\% & 93.5\% & 86.7\% & 95.0\% & 77.7\% & 73.0\% & 78.7\% \\
Mean & 90.2\% & 83.4\% & 93.7\% & 86.8\% & 95.3\% & 78.3\% & 73.1\% & 78.8\% \\
\bottomrule
\end{tabular}
\caption{The accuracy and $F_1$ score for five runs of the reproducibility variables selected to used for the large-scale conference paper analysis. }
\label{tab:reproducibility_variables_results_and_variation}
\end{table}

\begin{table}[h]
\centering
\begin{tabular}{c|c|c|c|c}
\toprule
\textbf{Provider} & \textbf{Model} & \shortstack{\textbf{Input Cost}\\\textbf{(1M Tokens)}} & \shortstack{\textbf{Output Cost}\\\textbf{(1M Tokens)}} & \shortstack{\textbf{Est. Input Cost}\\\textbf{(56\,800 Papers)}} \\
\midrule 
\multicolumn{5}{l}{\textit{Commercial Models}} \\
\midrule 
Anthropic & claude-opus-4-20250514 & \$15.00 & \$75.00 & \$11\,076.00 \\
Anthropic & claude-sonnet-4-20250514 & \$3.00 & \$15.00 & \$2\,215.20 \\
Anthropic & claude-3-5-haiku-20241022 & \$0.80 & \$4.00 & \$590.72 \\
\hline
Google & gemini-2.5-pro & \$1.25 & \$10.00 & \$923.00 \\
\textcolor{red}{\textbf{Google}} & \textcolor{red}{\textbf{gemini-2.5-flash}} & \textcolor{red}{\textbf{\$0.30}} & \textcolor{red}{\textbf{\$2.50}} & \textcolor{red}{\textbf{\$221.52}} \\
\textbf{Google} & \textbf{gemini-2.5-flash-lite} & \textbf{\$0.10} & \textbf{\$0.40} & \textbf{\$73.84} \\
\textbf{Google} &\textbf{ gemini-2.0-flash-001} & \textbf{\$0.15} & \textbf{\$0.60} & \textbf{\$110.76} \\
\hline
OpenAI & o3-2025-04-16 & \$2.00 & \$8.00 & \$1\,476.80 \\
OpenAI & o4-mini-2025-04-16 & \$1.10 & \$4.40 & \$812.24 \\
OpenAI & gpt-4.1-2025-04-14 & \$2.00 & \$8.00 & \$1\,476.80 \\
OpenAI & gpt-4o-2024-08-06 & \$2.50 & \$10.00 & \$1\,846.00 \\
\textbf{OpenAI} & \textbf{gpt-4.1-mini-2025-04-14} & \textbf{\$0.40} & \textbf{\$1.60} & \textbf{\$295.36} \\
\textbf{OpenAI} & \textbf{gpt-4.1-nano-2025-04-14} & \textbf{\$0.10} & \textbf{\$0.40} & \textbf{\$73.84} \\
\midrule 
\multicolumn{5}{l}{\textit{open weights Models}} \\
\midrule 
Google & gemma2:27B & - & - & - \\
Alibaba Cloud & qwen3:14B & - & - & - \\
Microsoft & phi4:14B & - & - & - \\
\bottomrule
\end{tabular}
\caption{The LLM models considered for automated paper analysis. The Anthropic models were excluded from final consideration due to the API not supporting structured output. The open weights models were excluded from final consideration due to the time it would take to complete the large-scale conference paper analysis with the hardware we had access to. The Google and OpenAI models that we estimated would cost more than \$300 USD for the Input Tokens for 56\,800 papers were excluded from final consideration due to cost. The bolded models were in the final consideration based on Accuracy. We selected gemini-2.5-flash (red) for the large-scale conference paper analysis.}
\label{tab:llm_models}
\end{table}

\begin{table}[h]
    \centering
    \begin{tabular}{l|rrrrrc}
    \toprule
        Variable & $F_1$-score & Precision & Recall & Accuracy & Std. Dev & 95\% CI \\
        \midrule
        Research Type & 100.00\% & 100.00\% & 100.00\% & 100.00\% & 0.00\% & [1.00, 1.00] \\
    Affiliation & 94.96\% & 92.96\% & 97.06\% & 95.62\% & 20.45\% & [0.92, 0.99] \\
    Pseudocode & 93.96\% & 88.61\% & 100.00\% & 94.38\% & 23.04\% & [0.91, 0.98] \\
    Open Code & 92.90\% & 88.89\% & 97.30\% & 93.12\% & 25.30\% & [0.89, 0.97] \\
    Open Source Data & 96.30\% & 94.35\% & 98.32\% & 94.38\% & 23.04\% & [0.91, 0.98] \\
    Dataset Splits & 81.48\% & 83.02\% & 80.00\% & 87.50\% & 33.07\% & [0.82, 0.93] \\
    Hardware Specification & 96.43\% & 94.74\% & 98.18\% & 97.50\% & 15.61\% & [0.95, 1.00] \\
    Software Dependencies & 81.25\% & 72.22\% & 92.86\% & 96.25\% & 19.00\% & [0.93, 0.99] \\
    Experiment Setup & 86.21\% & 76.34\% & 99.01\% & 80.00\% & 40.00\% & [0.74, 0.86] \\
    \bottomrule
    \end{tabular}
    \caption{Results on the evaluation dataset of our method per reproducibility variable. We present the $F_1$-score, precision, recall, and, accuracy with standard deviation and 95\% confidence interval.}
    \label{tab:test_set_results}
\end{table}

\clearpage

\section{Supplementary Figures}

\begin{figure}[!t]
    \centering
    \includegraphics[width=1\linewidth]{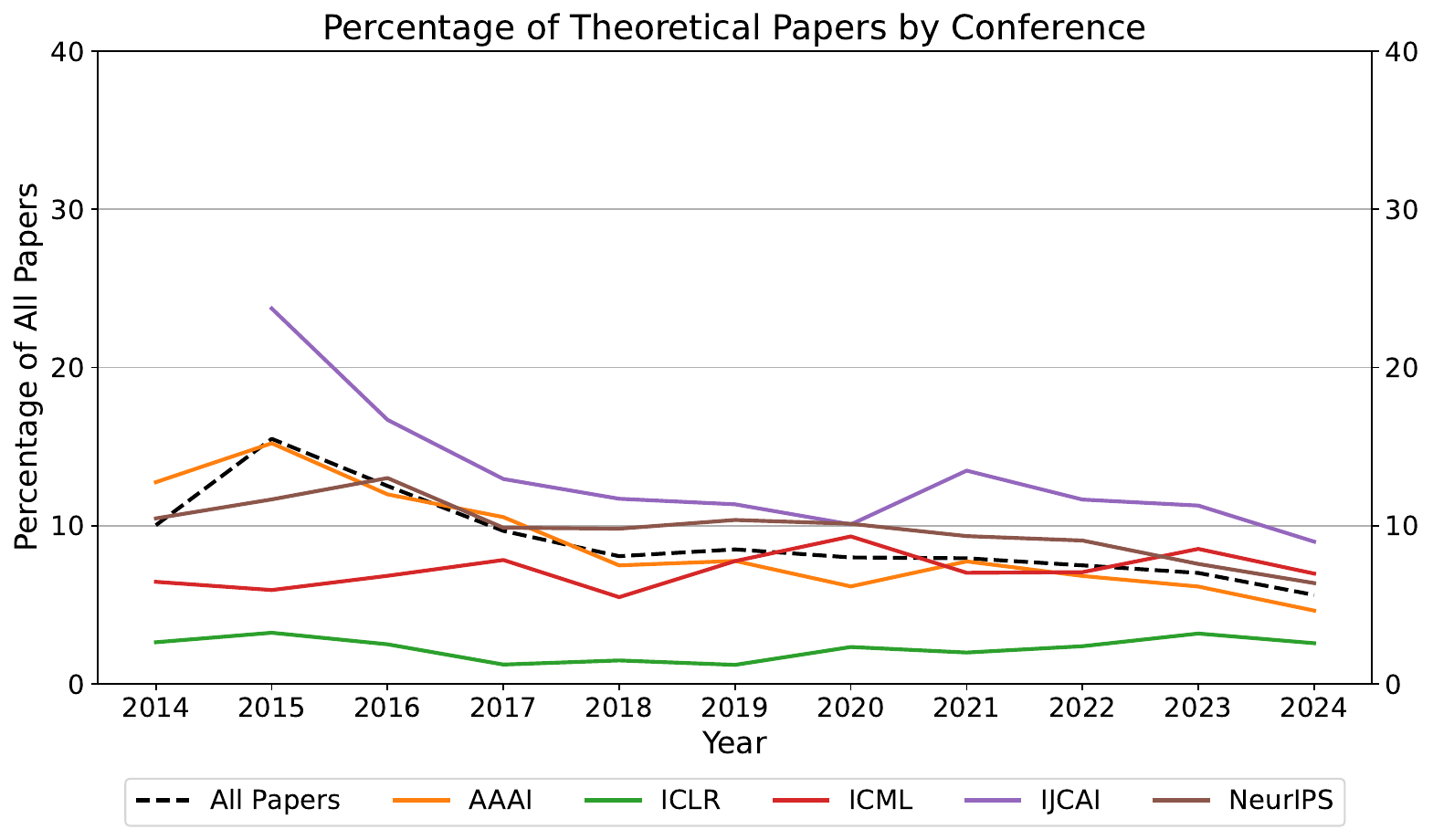}
    \caption{
    The percentage of theoretical papers published at the five AI conferences from 2014 to 2024. IJCAI did not hold a conference in 2014. Theoretical work declined from 10\% of all publications in 2014 to 6\% in 2024, following a peak of 15\% in 2015. This trend is most pronounced at IJCAI, which shifted from 24\% theoretical papers in 2015 to 9\% by 2024. The decline reflects a broader shift in AI research toward empirical methods, making open science practices, such as code and data sharing, increasingly critical for ensuring the reproducibility and trustworthiness of published findings. By 2024, empirical papers constituted 94\% of publications across these five conferences.}
    \label{figure:research_type}
\end{figure}

\begin{figure}[h]
    \centering
    \includegraphics[width=1\linewidth]{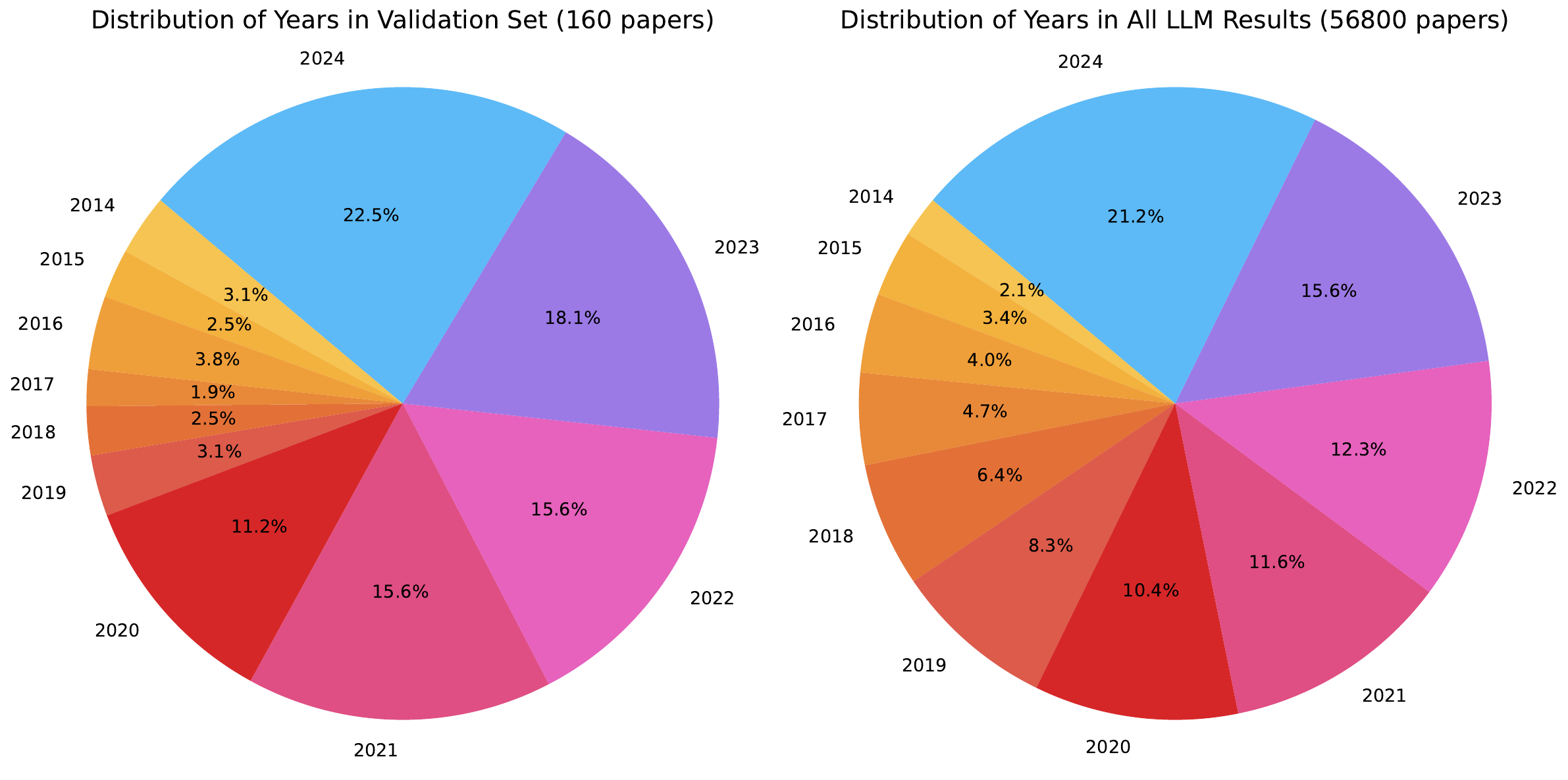}
    \caption{Year distributions of the 160-paper evaluation dataset (left) and the full 56,800-paper corpus (right). Each slice represents the proportion of papers from a given year. The distributions are broadly comparable, particularly from 2020 onward; papers from 2014–2019 are modestly overrepresented in the evaluation dataset relative to the full corpus, reflecting the smaller absolute paper counts in those years.}
    \label{figure:test_set_year_distribution}
\end{figure}

\begin{figure}[h]
    \centering
    \includegraphics[width=1\linewidth]{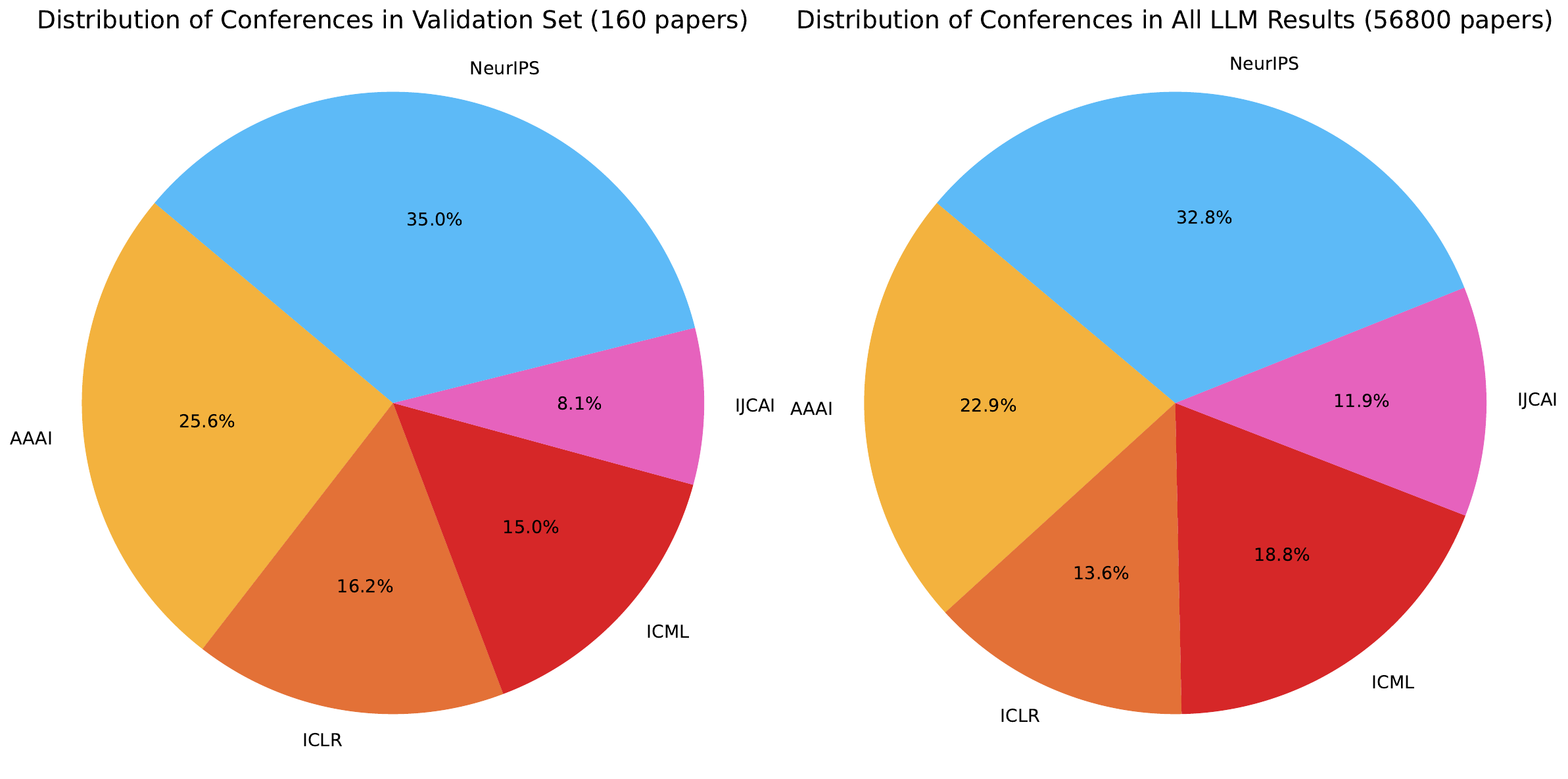}
    \caption{Conference distributions of the 160-paper evaluation dataset (left) and the full 56,800-paper corpus (right). Each slice represents the proportion of papers from a given conference. All five conferences are within three percentage points of their corresponding share in the full corpus, with the exception of ICML (15.0\% vs. 18.8\%) and IJCAI (8.1\% vs. 11.9\%), supporting the generalizability of the F1 score estimates reported in Table A10 across venues.}
    \label{figure:test_set_conference_distribution}
\end{figure}

\clearpage

\section{Paper Selection Methodology}
\label{sec:appendix-paper-selection-methodology}

The following section details the paper selection methodology for this work. We present the data sources, inclusion criteria, and exclusion criteria for each of the five leading AI conferences surveyed between 2014 and 2024.

\subsection*{AAAI Conference on Artificial Intelligence (AAAI)}
\begin{itemize}
    \item \textbf{Years Analysed:} 2014--2024
    \item \textbf{Data Source:} Paper metadata was collected from the official AAAI Conference Proceedings website: \url{https://aaai.org/aaai-publications/aaai-conference-proceedings/}.
    \item \textbf{Inclusion Criteria:} The analysis included all papers published in the main \textbf{Technical Tracks}.
    \item \textbf{Exclusion Criteria:} Papers from the following tracks were excluded: Student Abstract, Senior Member Summary Talks, Senior Member Blue Sky, Demonstrations, Doctoral Consortium, Senior Track, as well as papers from the affiliated EAAI and IAAI conferences.
    \item \textbf{Errors:} Two PDFs on AAAI's website from 2019 were corrupt and could not be read: "Deeply Fusing Reviews and Contents for Cold Start Users in Cross-Domain Recommendation Systems" and "TransNFCM: Translation-Based Neural Fashion Compatibility Modeling". The papers were downloaded from alternative locations, https://ojs.aaai.org/index.php/AAAI/article/view/3773/3651 and https://ojs.aaai.org/index.php/AAAI/article/view/3811/3689, respectively. 
\end{itemize}

\subsection*{International Conference on Learning Representations (ICLR)}
\begin{itemize}
    \item \textbf{Years Analysed:} 2014--2024
    \item \textbf{Data Sources:}
        \begin{itemize}
            \item \textbf{2014:} ICLR 2014 Archive (\url{https://iclr.cc/archive/2014/old-site/program-details/conference-program.html})
            \item \textbf{2015--2016:} ICLR Website Archives (\url{https://iclr.cc/archive/www/2015.html} \& \url{https://iclr.cc/archive/www/2016.html})
            \item \textbf{2017--2024:} OpenReview API (\url{https://openreview.net/group?id=ICLR.cc})
        \end{itemize}
    \item \textbf{Inclusion Criteria:}
        \begin{itemize}
            \item For 2014--2016, all papers designated as \textbf{Orals} and \textbf{Posters} were included.
            \item For 2017--2024, we included all submissions from the OpenReview API that did not have a ``reject'' status in the final decision field (the specific field name, such as \texttt{Acceptance\_Decision}, \texttt{Decision}, or \texttt{Meta\_Review}, varied by year). This encompassed papers published as \textbf{Orals}, \textbf{Posters}, and \textbf{Spotlights}.
        \end{itemize}
    \item \textbf{Exclusion Criteria:} All papers from \textbf{Workshop Tracks} were excluded across all years.
    \item \textbf{Errors:} The paper "The Concrete Distribution: A Continuous Relaxation of Discrete Random Variables" from 2017 returned an http 403 error. The paper was downloaded from an alternative location: \href{https://ora.ox.ac.uk/objects/uuid:5635eaad-c1bf-4f13-8832-034b90c0023a/files/m493237e01997411d8db562cb30b3ec71}{https://ora.ox.ac.uk/objects/uuid:5635eaad-c1bf-4f13-8832-034b90c0023a/files/m493237e01997411d8db562cb30b3ec71}. 
\end{itemize}

\subsection*{International Conference on Machine Learning (ICML)}
\begin{itemize}
    \item \textbf{Years Analysed:} 2014--2024
    \item \textbf{Data Source:} Proceedings of Machine Learning Research (PMLR) website: \url{https://proceedings.mlr.press}.
    \item \textbf{Inclusion Criteria:} The dataset includes \textbf{all papers} listed in the official proceedings for each year.
\end{itemize}

\subsection*{International Joint Conference on Artificial Intelligence (IJCAI)}
\begin{itemize}
    \item \textbf{Years Analysed:} 2015--2024
    \item \textbf{Data Source:} Official IJCAI proceedings website: \url{https://www.ijcai.org/all_proceedings}.
    \item \textbf{Inclusion Criteria:} Only papers from the main \textbf{Technical Track} were included.
    \item \textbf{Exclusion Criteria:} Papers from the following tracks were excluded: Demos, Journal Track, Best Papers from Sister Conferences, Doctoral Consortium, Invited Talks, Early Career, Survey track, Exhibits, AI and Arts, AI for Good, and Human-Centred AI.
\end{itemize}

\subsection*{Conference on Neural Information Processing Systems (NeurIPS)}
\begin{itemize}
    \item \textbf{Years Analysed:} 2014--2024
    \item \textbf{Data Source:} NeurIPS Proceedings website: \url{https://proceedings.neurips.cc}.
    \item \textbf{Inclusion Criteria:} All papers belonging to the \textbf{Main Conference Track} were included.
    \item \textbf{Exclusion Criteria:} Papers from the \textbf{Datasets and Benchmarks Track} were excluded from the analysis.
\end{itemize}

\begin{table}[h]
\centering

\label{tab:number_of_conference_papers}
\begin{tabular}{l|r|r|r|r|r|r}
\hline
\textbf{Year} & \textbf{AAAI} & \textbf{ICLR} & \textbf{ICML} & \textbf{IJCAI} & \textbf{NeurIPS} & \textbf{Total (Year)} \\
\hline
2014 & 447 & 38 & 310 & - & 411 & 1\,206 \\
2015 & 651 & 31 & 270 & 569 & 403 & 1\,924 \\
2016 & 676 & 80 & 322 & 647 & 569 & 2\,294\\
2017 & 645 & 245 & 434 & 664 & 679 & 2\,667 \\
2018 & 935 & 337 & 621 & 718 & 1\,009 & 3\,620 \\
2019 & 1\,146 & 502 & 773 & 846 & 1\,428 & 4\,695 \\
2020 & 1\,607 & 687 & 1\,084 & 645 & 1\,898 & 5\,921 \\
2021 & 1\,654 & 859 & 1\,183 & 586 & 2\,334 & 6\,616 \\
2022 & 1\,319 & 1\,094 & 1\,233 & 678 & 2\,671 & 6\,995 \\
2023 & 1\,578 & 1\,573 & 1\,828 & 639 & 3\,218 & 8\,836 \\
2024 & 2\,331 & 2\,260 & 2\,610 & 790 & 4\,035 & 12\,026 \\
\hline
\textbf{Total} & 12\,989 & 7\,706 & 10\,668 & 6\,782 & 18\,655 & 56\,800 \\
\end{tabular}
\caption{Number of Papers Analysed by AI Conferences and Year}
\end{table}

\begin{table}[h]
\centering
\begin{tabular}{lll}
\hline
\textbf{Conference} & \textbf{Year} & \textbf{URL} \\
\hline
NeurIPS & 2019 & \url{https://neurips.cc/Conferences/2019/CallForPapers} \\
AAAI    & 2021 & \url{https://aaai-23.aaai.org/reproducibility-checklist/} \\
IJCAI   & 2021 & \url{https://ijcai-21.org/wp-content/uploads/2020/12/20201226-IJCAI-Reproducibility.pdf} \\
ICML    & 2023 & \url{https://icml.cc/Conferences/2023/PaperGuidelines} \\
ICLR    & 2022$^{*}$ & \url{https://iclr.cc/Conferences/2022/AuthorGuide} \\
\hline
\end{tabular}
\caption{Reproducibility checklists and guidelines introduced by the five leading AI conferences. The year column indicates when each conference introduced its checklist or guideline. $^{*}$ICLR introduced an optional reproducibility statement guideline rather than a mandatory checklist.}
\label{tab:checklists-urls}
\end{table}

\section{LLM Prompt}
\label{sec:llm-prompt}

The prompt submitted to the model for each paper consisted of a single API query containing the full plain text of the paper followed by questions for each of the 20 reproducibility variables defined by ~\citet{gundersen_2018}.
Plain text was extracted from the PDF of each paper, excluding figures but including figure captions and table text; this extracted text was inserted into the prompt at the location indicated by the placeholder \textit{"PAPER TEXT INSERTED HERE"}.

Two JSON keys in the prompt differ from the variable names used in the main analysis. The train key corresponds to the open datasets variable, and the validation key corresponds to the dataset splits variable. 
These key names reflect the original variable definitions from ~\citet{gundersen_2018}, in which separate training, validation, and test data variables were defined. 
As described in \textbf{Section 3.1}, we replaced those three variables with open datasets and dataset splits to better capture artifact sharing practices relevant to reproducibility. 
The test variable from the original schema was retained in the prompt but excluded from the large-scale analysis.

The affiliation question in the prompt distinguishes three categories: academia (0), collaboration between academia and industry (1), and industry (2). 
As described in \textbf{Section 3.1}, the prompt optimization dataset contained only 11 purely industry-affiliated papers, which was insufficient to reliably assess classification performance for that category. 
We therefore combined industry and collaboration into a single binary category for the large-scale analysis, treating any paper with at least one industry-affiliated author as industry-affiliated.

\begin{lstlisting}[caption={Complete LLM Methodology Prompt}, label={lst:prompt}]
You will be provided with a research paper, and your task is to answer questions about the contents of the research paper.

Now, please analyze the following research paper:

<PAPER TEXT INSERTED HERE>

## Questions for Paper Analysis

Return the results of your analysis as a valid JSON object with the following structure. Each JSON key must contain two sub-elements:

### Output Format Requirements

Return your analysis as a valid JSON object with the following structure. Each field must contain two sub-elements:
- **result**: An Int with the specific value as defined for each question (varies by question - see below)
- **paper_text**: A string containing the relevant text excerpt from the paper that supports your answer, or a brief explanation if not found

The questions and JSON keys are as follows:


Determine whether the paper is based on **experimental research** or **theoretical research**.

Return **Experimental** (0) if:
- It conducts empirical studies, including running experiments, analyzing data, reporting metrics, or validating hypotheses
- Indicators include evaluation on datasets, comparisons to baselines, test/train/validation splits, ablation studies, and performance tables or graphs

Return **Theoretical** (1) if:
- It focuses only on conceptual or mathematical contributions without empirical validation
- Examples include algorithm design, proofs, lemmas, or symbolic derivations
- These papers may include pseudocode but do not analyze results from actual experiments

If the paper includes both theoretical and empirical components, classify it as **Experimental**.

Quote the text from the paper that supports your decision.

**Does the paper conduct EMPIRICAL STUDIES WITH DATA ANALYSIS (experiments, dataset evaluation, performance metrics, or hypothesis validation) rather than purely theoretical work?**
 Return 0 for experimental research or 1 for theoretical framework and use research_type as the JSON key.
Determine whether the paper reports a **positive** or **negative** outcome based on its main findings.

Return **Yes** (positive outcome) if:
- The authors explicitly state that their method improves upon prior work, achieves higher accuracy or F1 scores, or reaches state-of-the-art (SOTA) results
- Key indicators include phrases like "Our method significantly outperforms...", "We achieve better results than...", "These results confirm the effectiveness..."

Return **No** (negative outcome) if:
- The method fails to outperform baselines, results are marginal or inconclusive, or the hypothesis is not supported
- Typical phrases include "No significant improvement was observed...", "Comparable to existing methods...", "Results were not conclusive..."
- Only classify as positive if the paper clearly claims at least one success

Quote the text from the paper that supports your decision. If the answer is No, explain briefly why the information is insufficient.

**Does the paper report CLEAR SUCCESS CLAIMS (explicit statements of outperformance, higher accuracy, or confirmed effectiveness) in its main findings?**
 Return 1 for positive or 0 for negative and use result_outcome as the JSON key.
Determine whether the paper's authors have **industry**, **academic**, or **collaborative** affiliations.

The authors' affiliations and email addresses are usually listed in the first few paragraphs of the paper. Use this information to classify the affiliation type:

Return **2** (classify as Industry) if:
- All authors are affiliated with corporations or private-sector labs
- Email addresses end in domains like `.com`, `.ai`, `.tech`, or have company names like "IBM," "Google," "DeepMind," etc.

Return **0** (classify as Academia) if:
- All authors are from universities or public research institutions
- Email domains include `.edu`, `.ac.uk`, `.edu.cn`, or subdomains like `cs.cmu.edu`, `eecs.mit.edu`, etc.

Return **1** (classify as Collaboration) if:
- There is a mix of academic and industry affiliations based on the institutional names or email domains

Quote the text from the paper that supports your decision. If the answer is No, explain briefly why the information is insufficient.

**Does the paper provide CLEAR INSTITUTIONAL AFFILIATIONS (university names, company names, or email domains) that allow classification of author affiliation types?**
 Return 0 for academia only, 1 for collaboration of both academia and industry, or 2 for industry only and use affiliation as the JSON key.
Determine whether the paper includes an **explicit statement of the research problem**, using exact phrasing or close lexical variants.

Return **Yes** if:
- The paper includes a sentence that uses one of the following patterns: "the problem [is|we address|this paper addresses]...", "to solve [this problem|the problem of]..."
- Statements that clearly describe what the research aims to solve if they are framed like a problem, such as "the challenge we tackle is...", "the main issue is..."

Return **No** if:
- There are vague uses of the word "problem" or general task descriptions like "we study..." or "our method is designed to..." unless they directly align with the problem framing above

Quote the text from the paper that supports your decision. If the answer is No, explain briefly why the information is insufficient.

**Does the paper include an EXPLICIT RESEARCH PROBLEM STATEMENT (using phrases like "the problem is", "to solve the problem of", or "the challenge we tackle")?**
 Return 1 for yes or 0 for no and use problem_description as the JSON key.
Determine whether the paper includes an **explicit statement of the research objective or goal**, using exact phrasing or close lexical variants.

Return **Yes** if:
- The paper contains a sentence with one of these forms: "the objective [of this paper|is|was]...", "our objective [is|was]...", "the goal [of this research|is|was]...", "our goal [is|was]..."
- Clear variations such as "a key objective of this study is...", "one objective of this work is...", "the main goal of this approach is...", "this study has the objective/goal of...", "the research objective is defined as...", "an explicit goal is to..."

Return **No** if:
- There are vague motivation or general intent phrases (e.g., "we aim to...", "we explore...", "our focus is...") unless they use the words **goal** or **objective** in the sentence

Quote the text from the paper that supports your decision. If the answer is No, explain briefly why the information is insufficient.

**Does the paper explicitly state a RESEARCH GOAL OR OBJECTIVE (using the exact words "goal" or "objective" in formal research statements)?**
 Return 1 for yes or 0 for no and use goal_objective as the JSON key.
Determine whether the paper explicitly states the **research methodology**, using exact phrasing or close lexical variants.

Return **Yes** if:
- The paper contains a sentence that includes any of the following exact or near-exact expressions: "the methodology...", "our methodology...", "the research methodology...", "the research method...", "our research method..."
- Variants like "this paper uses the methodology...", "we follow the research method of..." are also acceptable, but only if they use the key phrases **"methodology"** or **"research method"** directly

Return **No** if:
- There are phrases like "our approach..." or "we propose a method..." unless they use one of the key phrases above

Quote the text from the paper that supports your decision. If the answer is No, explain briefly why the information is insufficient.

**Does the paper explicitly use the phrase METHODOLOGY OR RESEARCH METHOD (exact terminology "methodology" or "research method" in formal statements)?**
 Return 1 for yes or 0 for no and use research_method as the JSON key.
Determine whether the paper explicitly states one or more **research questions**, using exact phrasing or close lexical variants.

Return **Yes** if:
- The paper includes phrases like "the research questions are...", "this research seeks to answer...", "we pose the following research questions..."
- Lists labeled as "RQ", "RQ1", "RQ2", or "Research Question 1", etc.
- A clearly structured sentence like "we aim to answer the following question(s)...", followed by one or more question-like statements

Return **No** if:
- There are vague intent statements like "we explore..." or "we investigate..." unless they are tied to a numbered or explicitly labeled research question

Quote the text from the paper that supports your decision. If the answer is No, explain briefly why the information is insufficient.

**Does the paper clearly present FORMAL RESEARCH QUESTIONS (explicitly labeled as "RQ", "Research Question", or structured question statements)?**
 Return 1 for yes or 0 for no and use research_question as the JSON key.
Determine whether the paper explicitly states a **research hypothesis**, using exact phrasing or close lexical variants.

Return **Yes** if:
- The paper includes a sentence that uses the word **"hypothesis"** in any of the following accepted forms: "the hypothesis is...", "the hypothesis that...", "we hypothesize that...", "our hypothesis is...", "our hypothesis assumes...", "our guiding hypothesis is...", "the central hypothesis...", "this hypothesis suggests that...", "we test the hypothesis that...", "this paper tests the hypothesis...", "we formulate the hypothesis that...", references to a "Null Hypothesis" or "Alternative Hypothesis"

Return **No** if:
- There are statements that imply assumptions or beliefs unless the term **"hypothesis"** appears in the sentence

Quote the text from the paper that supports your decision. If the answer is No, explain briefly why the information is insufficient.

**Does the paper explicitly refer to a RESEARCH HYPOTHESIS (using the exact word "hypothesis" in formal research statements)?**
 Return 1 for yes or 0 for no and use hypothesis as the JSON key.
Determine whether the paper explicitly states a **predicted outcome**, using exact phrasing or close lexical variants.

Return **Yes** if:
- The paper contains a sentence that uses any of the following phrases: "we predict that...", "the predicted outcome is...", "we expect that...", "our expectation is that...", "it is expected that..."
- Phrasings like "the model is expected to..." or "our approach is predicted to..."

Return **No** if:
- There are vague goal statements like "we aim to..." or "we intend to..." unless they include the words **predict** or **expect**

Quote the text from the paper that supports your decision. If the answer is No, explain briefly why the information is insufficient.

**Does the paper include an EXPLICIT PREDICTED OUTCOME (using specific prediction language like "we predict", "we expect", or "predicted outcome")?**
 Return 1 for yes or 0 for no and use prediction as the JSON key.
Determine whether the paper explicitly states the **contributions** of the research, using exact phrasing or close lexical variants.

Return **Yes** if:
- The paper contains a sentence that clearly lists or describes contributions using any of the following: 
  - "The contributions of this paper are..."
  - "Our contributions are..."
  - "This paper contributes the following..."
  - "This contributes to..." (only if it refers to novel findings or technical contributions)
- Phrases like: "The main contribution is...", "A key contribution of this work is..."

Return **No** if:
- There are vague mentions of impact or relevance unless the word **"contribution"** is clearly tied to something new, original, or specific to this research

Quote the text from the paper that supports your decision. If the answer is No, explain briefly why the information is insufficient.

**Does the paper explicitly list or state the CONTRIBUTIONS using the accepted phrasing to describe novel findings or technical advances?**
 Return 1 for yes or 0 for no and use contribution as the JSON key.
Determine whether the paper contains **pseudocode** or a clearly labeled algorithm block.

Return **Yes** if:
- The paper includes a figure, block, or section labeled "Pseudocode", "Algorithm", or "Algorithm X"
- Structured steps for a method or procedure formatted like code or an algorithm (even if not explicitly called "pseudocode")

Return **No** if:
- The paper only describes steps in regular paragraph text without structured formatting

Quote the text from the paper that supports your decision. If the answer is No, explain briefly why the information is insufficient.

**Does the paper contain STRUCTURED PSEUDOCODE OR ALGORITHM BLOCKS (clearly labeled algorithm sections or code-like formatted procedures)?**
 Return 1 for yes or 0 for no and use pseudocode as the JSON key.
Determine whether the paper provides **open-source code for the methodology it describes**.

Return **Yes** if:
- The paper includes an unambiguous sentence where the authors state they are releasing the code **for the work described in this paper** (e.g., "We release our code...", "The source code for our method is available at...")
- A direct link to a source-code repository (e.g., GitHub, GitLab, Bitbucket) that contains the code for the paper's methodology
- A clear statement that code is provided in **supplementary material**, **appendices**, or via an **anonymous review link**

Return **No** if:
- The code is promised for the future using phrases like "we plan to release...", "code will be made available...", or "our tool will be publicly available..."
- The code is only available "upon request"
- The text discusses the source code of a **third-party tool or platform that the authors used**, but does not provide their own implementation code
- The link points to a resource that is explicitly a `dataset`, `benchmark`, `corpus`, `taxonomy`, or `data`, and does not also clearly host the source code
- The URL provided is for a general domain, a personal homepage, **a project demonstration page, or a high-level project overview page** instead of a specific code repository
- **The text is ambiguous or lacks a clear, affirmative statement of release**

Quote the text from the paper that supports your decision. If the answer is No, explain briefly why the information is insufficient.

**Does the paper provide CONCRETE ACCESS TO SOURCE CODE (specific repository link, explicit code release statement, or code in supplementary materials) for the methodology described in this paper?**
 Return 1 for yes or 0 for no and use open_source_code as the JSON key.
Determine whether the paper includes a link to the **actual code used to run the experiments** described in the paper.

Return **Yes** if:
- The paper provides a link to a GitHub, GitLab, Bitbucket, or anonymous repository and does **not** specify that the code is unrelated to experiments
- The text mentions that code, scripts, or software used to run the experiments is "available at...", "shared in supplementary material" or "appendix"
- The linked repo is implied to contain code that produces the **results, tables, or figures** in the paper

Return **No** if:
- The paper includes **no link**
- The link is unrelated to this paper's experiments (e.g., a third-party baseline, pretrained model, or unrelated library)
- The code is **only promised** or **available upon request**
- The link is to a **dataset**, **benchmark**, or **corpus**

Quote the text from the paper that supports your decision. If the answer is No, explain briefly why the information is insufficient.

**Does the paper include CONCRETE ACCESS TO EXPERIMENT CODE (specific repository links or explicit statements about code availability) used to run its experiments?**
 Return 1 for yes or 0 for no and use open_experiment_code as the JSON key.
Determine whether the paper explicitly states that the **dataset** used in the experiments is **publicly available** or an **open** dataset.

Return **Yes** if:
- The paper uses a well-known public dataset (e.g., "CIFAR-10", "MNIST", "ImageNet", "COCO", "Penn Treebank")
- Provides a direct URL, DOI, or specific repository name (e.g., GitHub, Zenodo, Figshare) where the dataset can be accessed
- Cites a published paper or resource that contains the dataset with proper bibliographic information **including author names and year in brackets or parentheses**
- States the dataset is in supplementary material with specific file names or section references
- **References standard academic datasets with citations** or **mentions datasets from well-known repositories or benchmarks with proper attribution**

Return **No** if:
- The paper mentions the dataset but gives no indication of availability
- The dataset is proprietary, private, or internal
- The authors created their own dataset but do not provide public access (no link, DOI, repository, or citation)
- The paper describes or mentions a dataset but does not provide any source, link, citation, or repository information for accessing it
- The dataset is described as "available upon request" or "available from authors" without permanent public access
- The paper only describes the dataset characteristics, collection process, or statistics without providing access information
- The paper mentions using "publicly available data" but does not specify the exact source or provide access details
- **The dataset name is mentioned but no citation, link, repository, or author attribution is provided**

Quote the text from the paper that supports your decision. If the answer is No, explain briefly why the information is insufficient.

**Does the paper provide CONCRETE ACCESS INFORMATION (specific link, DOI, repository name, formal citation with authors/year, or reference to established benchmark datasets) for a publicly available or open dataset?**
 Return 1 for yes or 0 for no and use train as the JSON key.
Determine whether the paper explicitly provides **training/test/validation dataset splits** needed to reproduce the experiment.

Return **Yes** if:
- The paper specifies exact split percentages (e.g., "80/10/10 split", "70% training, 15% validation, 15% test")
- Provides absolute sample counts for each split (e.g., "40,000 training samples, 5,000 validation, 5,000 test")
- References predefined splits with citations (e.g., "we use the standard train/test split from [Author et al., 2020]")
- Mentions specific file names or URLs for custom splits (e.g., "train.txt, val.txt, test.txt available at...")
- Describes stratified or group-based splitting methodology (e.g., "stratified by class", "split by subject ID", "temporal split")
- Provides random seed with splitting strategy (e.g., "random split with seed 42")
- Specifies cross-validation setup (e.g., "5-fold cross-validation", "leave-one-out cross-validation")
- Uses standard benchmark splits that are well-defined (e.g., "CIFAR-10 standard split", "ImageNet validation set")

Return **No** if:
- References datasets without mentioning splits (e.g., "we use the XYZ dataset")
- Splits are mentioned vaguely (e.g., "we split the data appropriately")
- Only mentions total dataset size without split information
- Defers split details to supplementary materials or other papers without providing access
- Uses phrases like "standard split" or "typical split" without specification

Quote the text from the paper that supports your decision. If the answer is No, explain briefly why the information is insufficient.

**Does the paper provide SPECIFIC DATASET SPLIT INFORMATION (exact percentages, sample counts, citations to predefined splits, or detailed splitting methodology) needed to reproduce the data partitioning?**
 Return 1 for yes or 0 for no and use validation as the JSON key.
Determine whether the paper explicitly states the **test data split** or **testing subset** used to evaluate the model's performance.

Return **Yes** if:
- The paper explicitly mentions phrases like "test set", "testing set", "test portion", "test split", "held-out test data"
- Mentions of dataset names used for testing (e.g., "CIFAR-10 test set", "evaluated on the test set of ImageNet")
- Quantitative indicators such as "10K examples for testing", "20% used as test data"

Return **No** if:
- There are vague mentions like "test accuracy" or "we tested our model" with no details about the test data
- "Test results" or "testing performance" without indicating the data used
- References to standard datasets (like "we used MNIST") without specifying the test subset or split

Quote the text from the paper that supports your decision. If the answer is No, explain briefly why the information is insufficient.

**Does the paper clearly describe the TEST DATA SPLIT (explicit mention of test set, split, or subset with specific details) used for evaluation?**
 Return 1 for yes or 0 for no and use test as the JSON key.
Determine whether the paper includes links to the **results of the experiment**.

Return **Yes** if:
- The paper includes a URL (e.g., to GitHub, Google Drive, institutional repository, Zenodo, Dropbox, etc.) pointing to result files, visualizations, spreadsheets, or dashboards
- It says results are in supplementary material, appendix, or provided anonymously for review
- It includes phrases like "Results available at...", "Our results can be found in...", "The experimental output is linked at..."

Return **No** if:
- Results are only shown in the paper's main text (figures or tables inside the PDF)
- The paper mentions results will be released later or are available upon request
- The link is only to datasets, tools, or pretrained models not tied to the paper's experimental output

Quote the text from the paper that supports your decision. If the answer is No, explain briefly why the information is insufficient.

**Does the paper link to or externally share CONCRETE ACCESS TO EXPERIMENTAL RESULTS (specific URLs, repositories, or supplementary materials containing result files)?**
 Return 1 for yes or 0 for no and use results as the JSON key.
Determine whether the paper **explicitly describes the hardware used** to run its experiments.

Return **Yes** if:
- The paper mentions any specific hardware setup, including:
  - **Specific GPU models** (e.g., "NVIDIA A100", "RTX 2080 Ti", "Tesla V100")
  - **Specific CPU models or processors** (e.g., "Intel Core i7-4770K", "Intel Xeon E5-2630", "AMD Ryzen 9 5950X", "Intel Core i5-3210M", "Intel i7-2600")
  - **TPU or other accelerator references** (e.g., "TPU v2", "Google TPU", "Intel Neural Compute Stick")
  - **Cloud or cluster resources with specs** (e.g., "AWS p3.8xlarge with V100 GPUs", "Google Cloud TPU v3")
  - **Computer specifications with processor details** (e.g., "laptop with Intel Core i5", "workstation with Intel Xeon", "desktop PC with Intel Core i7")
- Direct statements about where experiments were run, such as:
  - "We trained our models using..."
  - "Experiments were performed on..."
  - "Hardware specifications include..."
  - "All experiments ran on..."
  - "Our machine has..."
  - "Results are obtained on..."

Return **No** if:
- No hardware is mentioned
- The paper only discusses software, datasets, or **vague terms like "on a GPU", "using a High Performance Computing Resource", or "on a server" without any specific model numbers, processor types, or memory details**
- Any mention of hardware is disconnected from the experimental process
- The hardware mentioned (like mobile phones, tablets, or IoT devices) is not used for training or inference in the experiments
- **Only general computing environments are mentioned without any specific hardware details (e.g., "on a cluster", "using cloud computing" with no specifications)**

Quote the text from the paper that supports your decision. If the answer is No, explain briefly why the information is insufficient.

**Does the paper provide SPECIFIC HARDWARE DETAILS (exact GPU/CPU models, processor types with speeds, memory amounts, or detailed computer specifications) used for running its experiments?**
 Return 1 for yes or 0 for no and use hardware_specification as the JSON key.
Determine whether the paper provides a reproducible description of the ancillary software. A reproducible description **must include specific version numbers** for key software components.

Return **Yes** if the paper meets one of the following criteria:
- It lists multiple key software components with their versions (e.g., "Python 3.8, PyTorch 1.9, and CUDA 11.1")
- It names a self-contained solver, simulation environment, or specialized package with a specific version number (e.g., "CPLEX 12.4", "Gecode 4.2.0", "Choco 2.1.5")

Return **No** if the paper only mentions:
- Software names without version numbers (e.g., "using Caffe", "the scikit-learn package")
- A programming language, even with a version, without listing any versioned libraries or solvers (e.g., "implemented in Java 7" by itself is not enough)

Quote the text from the paper that supports your decision. If the answer is No, explain briefly why the information is insufficient.

**Does the paper provide SPECIFIC ANCILLARY SOFTWARE DETAILS (e.g., library or solver names with version numbers like Python 3.8, CPLEX 12.4) needed to replicate the experiment?**
 Return 1 for yes or 0 for no and use software_dependencies as the JSON key.
Determine whether the paper explicitly provides details about the **experimental setup**, especially hyperparameters or system-level training settings.

Return **Yes** if:
- The paper contains specific hyperparameter values (e.g., learning rate, batch size, number of epochs, optimizer settings)
- Details on model initialization, dropout rate, or training schedules
- A clearly labeled table or paragraph describing training settings
- A section titled "Experimental Setup" or similar with configuration information

Return **No** if:
- The paper only mentions that "we trained a model" or refers to "standard settings" without elaboration
- The only details about training are deferred to supplemental materials, code, or prior work
- There is no mention of hyperparameters, optimizer settings, or explicit configuration steps

Quote the text from the paper that supports your decision. If the answer is No, explain briefly why the information is insufficient.

**Does the paper contain SPECIFIC EXPERIMENTAL SETUP DETAILS (concrete hyperparameter values, training configurations, or system-level settings) in the main text?**
 Return 1 for yes or 0 for no and use experiment_setup as the JSON key.

\end{lstlisting}

\section{Selected Results}

\autoref{tab:selected-results} presents a sample of 10 papers drawn from the 160-paper evaluation dataset, selected to represent both the temporal and venue breadth of the full corpus: the sample includes papers from most years covered by the study and from each of the five conferences. 
For each paper, the table reports the ground truth label assigned by a human annotator, the LLM classification, and the supporting text extracted by the LLM in response to the instruction "Quote the text from the paper that supports your decision." Rows in which the LLM classification disagrees with the ground truth are shown in bold.

The papers were additionally selected to include at least one misclassification for each of the nine reproducibility variables included in the large-scale analysis, with the exception of research type, for which the LLM achieved 100\% accuracy on the evaluation dataset. 
Author names in the supporting text for the affiliation variable have been replaced with *** to remove identifying information.

\newgeometry{left=0.75in,right=0.75in,top=1in,bottom=1in}
\begin{landscape}

\setlength{\LTcapwidth}{\linewidth}

\begin{longtable}{|p{0.18\linewidth}|p{0.1\linewidth}|p{0.1\linewidth}|p{0.54\linewidth}|}
\hline
\textbf{Reproducibility Variable} & \textbf{Manual Label} & \textbf{LLM Label} & \textbf{LLM Supporting Text} \\
\hline
\endfirsthead
\hline
\textbf{Reproducibility Variable} & \textbf{Manual Label} & \textbf{LLM Label} & \textbf{LLM Supporting Text} \\
\hline
\endhead
\hline
\endfoot
\multicolumn{4}{|l|}{\textbf{ICLR 2024 - How Connectivity Structure Shapes Rich and Lazy Learning in Neural Circuits}} \\
\hline
Research Type & Empirical & Empirical & Through both empirical and theoretical analyses, we discover that high-rank initializations typically yield smaller network changes indicative of lazier learning, a finding we also confirm with experimentally-driven initial connectivity in recurrent neural networks. \\
\hline
Affiliation & Industry & Industry & 1University of Washington, Seattle, WA, USA 2Allen Institute for Brain Science, Seattle WA, USA 3Mila Quebec AI Institute, Montreal, QC, Canada 4Samsung SAIT AI Lab, Montreal, QC, Canada 5Mc Gill University, Montreal, QC, Canada 6Canada CIFAR AI Chair, CIFAR, Toronto, ON, Canada 7Universit\'e de Montr\'eal, Montreal, QC, Canada \\
\hline
Pseudocode & False & False & The paper does not contain any structured pseudocode or algorithm blocks. \\
\hline
Open Source Code & False & False & The paper does not provide any explicit statements about releasing source code or links to a code repository. \\
\hline
Open Datasets & True & True & "For our investigations, we applied this initialization scheme across a variety of cognitive tasks including two-alternative forced choice (2AF), delayed-match-to-sample (DMS), context-dependent decision-making (CXT) tasks implemented with Neurogym (Molano-Mazon et al., 2022) and the well-known machine learning benchmark sequential MNIST (s MNIST). \\
\hline
Dataset Splits & False & False & The paper does not provide specific dataset split information (exact percentages, sample counts, or detailed splitting methodology) for reproduction. \\
\hline
Hardware Specification & False & False & The paper does not provide specific hardware details (e.g., GPU/CPU models, processor types, or memory amounts) used for running its experiments. \\
\hline
Software Dependencies & False & False & The paper mentions general tools like PyTorch in its references, but it does not specify any software dependencies with version numbers used for its experimental setup. \\
\hline
Experiment Setup & False & False & The paper states 'Details of parameter settings can be found in Appendix C.' and refers to Appendix C for other setup specifics, but these details are not provided in the main text. \\
\hline
\multicolumn{4}{|l|}{\textbf{NeurIPS 2015 - Online Gradient Boosting}} \\
\hline
Research Type & Empirical & Empirical & Finally, we conduct some proof-of-concept experiments which show that our online boosting algorithms do obtain performance improvements over di erent classes of base learners. All experiments were done on a collection of 14 publically available regression and classification datasets (described in Section D in the supplementary material) using squared loss. The following table reports the average and the median, over the datasets, relative improvement in squared loss over the respective base learner. \\
\hline
Affiliation & Industry & Industry & *** Yahoo Labs New York, NY 10036 ***@yahoo-inc.com *** Princeton University Princeton, NJ 08540 ***@cs.princeton.edu *** Yahoo Labs New York, NY 10036 ***@yahoo-inc.com *** Princeton University Princeton, NJ 08540 haipengl@cs.princeton.edu \\
\hline
Pseudocode & True & True & Algorithm 1 Online Gradient Boosting for span(F) Algorithm 2 Online Gradient Boosting for CH(F) \\
\hline
Open Source Code & True & True & Is it possible to boost in an online fashion in practice with real base learners? To study this question, we implemented and evaluated Algorithms 1 and 2 within the Vowpal Wabbit (VW) open source machine learning system [23]. [23] VW. URL https://github.com/John Langford/vowpal\_wabbit/. The paper states that the algorithms were implemented *within* Vowpal Wabbit (an existing open-source system), but it does not state that the code for *their specific implementations* of Algorithms 1 and 2 is open-source or provided. \\
\hline
\textbf{Open Datasets} & \textbf{False} & \textbf{True} & \textbf{All experiments were done on a collection of 14 publically available regression and classification datasets (described in Section D in the supplementary material) using squared loss.} \\
\hline
\textbf{Dataset Splits} & \textbf{False} & \textbf{True} & \textbf{Parameters were tuned based on progressive validation loss on half of the dataset; reported is propressive validation loss on the remaining half. Progressive validation is a standard online validation technique, where each training example is used for testing before it is used for updating the model [3].} \\
\hline
Hardware Specification & False & False & The paper does not provide any specific hardware details such as GPU/CPU models, memory, or cloud instance types used for running the experiments. It only states that experiments were conducted. \\
\hline
Software Dependencies & False & False & The paper states that the algorithms were 'implemented and evaluated [...] within the Vowpal Wabbit (VW) open source machine learning system [23]', but it does not specify a version number for Vowpal Wabbit or any other software dependencies. \\
\hline
\textbf{Experiment Setup} & \textbf{False} & \textbf{True} & \textbf{The only parameters tuned were the learning rate and the number of weak learners, as well as the step size parameter for Algorithm 1.} \\
\hline
\multicolumn{4}{|l|}{\textbf{ICLR 2023 - Offline Reinforcement Learning with Differentiable Function Approximation is Provably Efficient}} \\
\hline
Research Type & Theoretical & Theoretical & We show ofﬂine RL with differentiable function approximation is provably efficient by analyzing the pessimistic fitted Q-learning (PFQL) algorithm, and our results provide the theoretical basis for understanding a variety of practical heuristics that rely on Fitted Q-Iteration style design. In addition, we further improve our guarantee with a tighter instance-dependent characterization. We hope our work could draw interest in studying reinforcement learning with differentiable function approximation beyond the scope of current research. \\
\hline
Affiliation & Academic  & Academic  & Department of Computer Science Department of Electrical and Computer Engineering University of California, Santa Barbara Princeton University \\
\hline
Pseudocode & True & True & Algorithm 1 Pessimistic Fitted Q-Learning (PFQL) Algorithm 2 Vanilla Fitted Q-Learning (VFQL) Algorithm 3 Variance-Aware Fitted Q Learning (VAFQL). \\
\hline
Open Source Code & False & False & The paper does not provide any statement about making its source code available or links to a code repository. \\
\hline
Open Datasets & False & False & This is a theoretical paper and does not describe empirical experiments or the use of any datasets for training. \\
\hline
Dataset Splits & False & False & This is a theoretical paper and does not describe empirical experiments or data splits for training, validation, or testing. \\
\hline
Hardware Specification & False & False & This is a theoretical paper and does not describe empirical experiments, therefore, it does not mention any hardware specifications. \\
\hline
Software Dependencies & False & False & This is a theoretical paper and does not describe empirical experiments, therefore, it does not list any software dependencies with specific version numbers. \\
\hline
Experiment Setup & False & False & This is a theoretical paper and does not describe empirical experiments, therefore, it does not provide details about an experimental setup or hyperparameters. \\
\hline
\multicolumn{4}{|l|}{\textbf{NeurIPS 2022 - Look where you look! Saliency-guided Q-networks for generalization in visual Reinforcement Learning}} \\
\hline
Research Type & Empirical & Empirical & SGQN vastly improves the generalization capability of Soft Actor-Critic agents and outperforms existing state-of-the-art methods on the Deepmind Control Generalization benchmark, setting a new reference in terms of training efficiency, generalization gap, and policy interpretability. \\
\hline
\textbf{Affiliation} & \textbf{Academic} & \textbf{Industry} & \textbf{***IRT Saint-Exup\'ery ISAE-SUPAERO IMT, INSA Toulouse ANITI Toulouse, France ***@irt-saintexupery.com *** IRT Saint-Exup\'ery ISAE-SUPAERO Toulouse, France **@irt-saintexupery.com *** IRT Saint-Exup\'ery IMT, Universit\'e Paul Sabatier Toulouse, France ***@irt-saintexupery.com*** ISAE-SUPAERO Universit\'e de Toulouse ANITI Toulouse, France ***@isae-supaero.fr} \\
\hline
Pseudocode & True & True & Algorithm 1 presents the pseudo-code of combining SGQN with SAC, yielding an SG-SAC algorithm. \\
\hline
\textbf{Open Source Code} & \textbf{False} & \textbf{True} & \textbf{Did you include the code, data, and instructions needed to reproduce the main experimental results (either in the supplemental material or as a URL)? [Yes]} \\
\hline
Open Datasets & True & True & We compare our approach with current state-of-the-art methods for generalization in continuous actions RL (RAD [Laskin et al., 2020b], Dr Q [Yarats et al., 2020], SODA [Hansen and Wang, 2021], SVEA [Hansen et al., 2021]) on five environments from the DMControl Generalization benchmark (DMControl-GB) [Hansen and Wang, 2021]. \\
\hline
Dataset Splits & False & False & The paper discusses training on a "vanilla training environment" and evaluating on "video easy" and "video hard" benchmarks, but does not explicitly mention a "validation" split for purposes like hyperparameter tuning. \\
\hline
\textbf{Hardware Specification} & \textbf{False} & \textbf{True} & \textbf{This work benefited from computing resources from CALMIP under grant P21001.} \\
\hline
Software Dependencies & False & False & The paper does not explicitly provide specific software dependencies with version numbers in the main text or the provided ethical checklist. \\
\hline
\textbf{Experiment Setup} & \textbf{False} & \textbf{True} & \textbf{We trained all agents for 500 000 steps using the vanilla training environment with no visual variation. Appendix A, B, and E discuss all the hyperparameters, network architectures, and implementation choices used for this benchmark.} \\
\hline
\multicolumn{4}{|l|}{\textbf{IJCAI 2022 - Parameterized Complexity of Hotelling-Downs with Party Nominees}} \\
\hline
Research Type & Theoretical & Theoretical & We provide FPT and XP algorithms and we complement them with a W[1]-hardness result. \\
\hline
Affiliation & Academic & Academic & ***, ***, *** Royal Holloway, University of London \{***, ***\}@rhul.ac.uk, ***@gmail.com \\
\hline
Pseudocode & False & False & The paper describes algorithmic approaches, such as dynamic programming, in prose, but it does not include any explicitly labeled 'Pseudocode' or 'Algorithm' blocks, nor does it present structured steps formatted like code. \\
\hline
Open Source Code & False & False & The paper is theoretical and does not mention providing any open-source code for the described methodologies or models. \\
\hline
Open Datasets & False & False & This is a theoretical paper and does not involve the use or release of any public datasets for training. \\
\hline
Dataset Splits & False & False & This is a theoretical paper and does not involve dataset splits for validation or any other purpose. \\
\hline
Hardware Specification & False & False & The paper is theoretical and does not describe any experimental setup that would require hardware specifications. \\
\hline
Software Dependencies & False & False & The paper is theoretical and does not specify any software dependencies with version numbers for implementation or experimentation. \\
\hline
Experiment Setup & False & False & The paper is theoretical and does not present experimental results, therefore no experimental setup details like hyperparameters or training configurations are provided. \\
\hline
\multicolumn{4}{|l|}{\textbf{AAAI 2021 - Escaping Local Optima with Non-Elitist Evolutionary Algorithms}} \\
\hline
Research Type & Empirical & Empirical & The theoretical analysis is complemented with an empirical investigation on instances of the set cover problem, showing that non-elitist EAs can perform better than the elitist ones. We also provide examples where usage of mutation rates close to the error thresholds is beneficial when employing non-elitist population-based EAs \\
\hline
Affiliation & Academic & Academic & 1Southampton Business School, University of Southampton, University Road, Southampton, SO17 1BJ, United Kingdom 2Sobolev Institute of Mathematics SB RAS, 13, Pevtsov str., Omsk, 644099, Russia 3School of Computer Science, University of Birmingham, Edgbaston, B15 2TT Birmingham, United Kingdom" \\
\hline
Pseudocode & True & True & Algorithm 1 (Dang and Lehre 2016a) Require: Initial population P0 X $\lambda$, parameter $\chi$ [0, n]. 1: for t N until a termination cond. is met do 2: for i = 1 to $\lambda$ do 3: Sample It(i) psel(Pt), and set x := Pt(It(i)). 4: Sample x pmut(x, $\chi$), and set Pt+1(i) := x . 5: end for 6: end for \\
\hline
\textbf{Open Source Code} & \textbf{False} & \textbf{True} &\textbf{ The plots with confidence intervals, and the source code with the instruction on how to reproduce the experiments are provided in the supplementary material.} \\
\hline
Open Datasets & True & True & We carried out some preliminary experiments with the non-elitist EAs on several instances of the Set Cover Problem (SCP) from the OR-library (Beasley 1990), representing the CYC and CLR families (Grossman and Wool 1997) and the Stein (Fulkerson, Nemhauser, and Trotter 1974) family of hard unicost benchmarks. \\
\hline
Dataset Splits & False & False & The paper mentions using instances of the Set Cover Problem but does not provide specific details on training, validation, or test dataset splits, such as percentages or sample counts. \\
\hline
Hardware Specification & True & True & They were compiled, then called from a Python program running on a server machine with AMD EPYC 7502 processors, Ubuntu 20.04 OS, GCC 9.3.0 and Python 3.8.3. \\
\hline
Software Dependencies & True & True & They were compiled, then called from a Python program running on a server machine with AMD EPYC 7502 processors, Ubuntu 20.04 OS, GCC 9.3.0 and Python 3.8.3. \\
\hline
Experiment Setup & True & True & Each algorithm was allowed the same budget of 2 108 function calls to evaluate its solutions, and each setting on each instance were tested with 40 replications of the run using different random seeds. The algorithms were implemented in C++ using its standard library for the random number generation. The graphs of non-elitist EAs with tournament selection are passing close to the best-known solution size 144 when $\chi$ = 0.67 if the tournament size is k = 2, and when $\chi$ = 1.08 if k = 3. \\
\hline
\multicolumn{4}{|l|}{\textbf{ICML 2023 - Towards Practical Preferential Bayesian Optimization with Skew Gaussian Processes}} \\
\hline
Research Type & Empirical & Empirical & towards building a more practical preferential BO, we develop a new method that achieves both high computational efficiency and low sample complexity, and then demonstrate its effectiveness through extensive numerical experiments.Numerical experiments on 12 benchmark functions show that the proposed method achieves better or at least competitive performance in both terms of computational efficiency and sample complexity compared with Gaussian approximation-based preferential BO (Gonz alez et al., 2017; Siivola et al., 2021; Fauvel \& Chalk, 2021) and MCMC-based preferential BO (Benavoli et al., 2021a;b), respectively. \\
\hline
Affiliation & Industry & Industry & *** 1 2 3 *** 2 *** 1 n1Nagoya Institute of Technology, Aichi, Japan 2Cyber Agent, Tokyo, Japan 3RIKEN AIP, Tokyo, Japan. \\
\hline
Pseudocode & True & True & Algorithm 1 shows the procedure of HB.Algorithm 2 shows the pseudo-code. \\
\hline
Open Source Code & True & True & Our experimental codes are publicly available at https: //github.com/Cyber Agent AILab/preferential BO. \\
\hline
Open Datasets & True & True & We employed the 12 benchmark functions.All the details of benchmark functions are shown in https://www.sfu.ca/ ssurjano/optimization.html. \\
\hline
Dataset Splits & False & False & The paper evaluates the optimization algorithms on benchmark functions. It mentions "10 random initialization" for regret calculation but does not specify explicit train/validation/test dataset splits in the conventional sense for supervised learning tasks. \\
\hline
Hardware Specification & False & False & The paper does not provide specific hardware details such as GPU or CPU models used for running the experiments. \\
\hline
Software Dependencies & False & False & The paper mentions general software components like 'RBF kernel' and 'Python', and refers to an implementation by other authors, but does not provide specific version numbers for key software dependencies or libraries. \\
\hline
\textbf{Experiment Setup} & \textbf{False} & \textbf{True} & \textbf{For preferential GP models, we use RBF kernel with automatic relevance determination (Rasmussen \& Williams, 2005), whose lengthscales are selected by marginal likelihood maximization per 10 iterations, and set fixed noise variance $\sigma$2 noise = 10 4.For the parameters for Gibbs sampling, burn-in is 1000, and the MC sample size for Duel UCB and EIIG is 1000 (thinning is not performed).For HB-UCB, we use $\beta$1/2 = 2.} \\
\hline
\multicolumn{4}{|l|}{\textbf{NeurIPS 2024 - Voila-A: Aligning Vision-Language Models with User's Gaze Attention}} \\
\hline
Research Type & Empirical & Empirical & We evaluate Voila-A using a hold-out validation set and a newly collected VOILA-GAZE test set, which features real-life scenarios captured with a gaze-tracking device. Our experimental results demonstrate that Voila-A significantly outperforms several baseline models. \\
\hline
Affiliation & Industry & Industry & ***1 , ***2 , ***3, ***2, ***3, ***1 1 SKLSDE Lab, Beihang University 2 Key Laboratory of Pervasive Computing, Tsinghua University 3 Microsoft Research \\
\hline
Pseudocode & False & False & The paper does not contain any clearly labeled 'Pseudocode' or 'Algorithm' blocks, nor does it present structured steps in a pseudocode-like format. \\
\hline
Open Source Code & True & True & Our code is available at https://github.com/naykun/Voila-A \\
\hline
Open Datasets & True & True & Table 1: Statistics of Voila-COCO and Voila-Gaze Datasets, SR refers to Survival Rate from raw data after filtering DATASET SPLIT \#IMAGES \#QUESTIONS SR VOILA-COCO TRAINING 20000 70000 93.5\% VOILA-COCO VALIDATION 100 550 71.1\% VOILA-COCO TEST 500 1900 75.7\% \\
\hline
Dataset Splits & True & True & Table 1: Statistics of Voila-COCO and Voila-Gaze Datasets, SR refers to Survival Rate from raw data after filtering DATASET SPLIT \#IMAGES \#QUESTIONS SR VOILA-COCO TRAINING 20000 70000 93.5\% VOILA-COCO VALIDATION 100 550 71.1\% VOILA-COCO TEST 500 1900 75.7\% \\
\hline
Hardware Specification & False & False & The paper mentions training models and optimizing, but does not specify the particular GPU or CPU models, memory, or other hardware components used for the experiments. Section F, which is cited for compute resources in the NeurIPS checklist, only details model configurations and training parameters, not hardware. \\
\hline
\textbf{Software Dependencies} & \textbf{False} & \textbf{True} & \textbf{The text model is an instance of MPTFor Causal LM 7B and the vision model is based on the CLIP Vi T-L/14 [42] vision encoder. ... The tokenizer used is Eleuther AI/gpt-neox-20b. The model s torch data type is set to bfloat16.} \\
\hline
Experiment Setup & True & True & For optimization, we employ the Adam W optimizer [22] with a starting learning rate of 1e-5 and a batch size of 4. We train Voila for three epochs, scheduling the learning rate using a cosine annealing scheduler. To prevent exploding gradients, we apply gradient clipping with a threshold of 1.0. \\
\hline
\multicolumn{4}{|l|}{\textbf{ICML 2021 - Order-Agnostic Cross Entropy for Non-Autoregressive Machine Translation}} \\
\hline
Research Type & Empirical & Empirical & Extensive experiments on major WMT benchmarks show that OAXE substantially improves translation performance, setting new state of the art for fully NAT models. \\
\hline
Affiliation & Industry & Industry & 1School of Computing and Information System, Singapore Management University, Singapore. Work was done when Cunxiao Du was under the Rhino-Bird Elite Training Program of Tencent AI Lab. 2Tencent AI Lab, China. \\
\hline
\textbf{Pseudocode} & \textbf{False} & \textbf{True} & \textbf{We use Hungarian algorithm to efficiently implement OAXE (e.g., 7 lines of core code, see Appendix A.1)} \\
\hline
Open Source Code & True & True & Our code, data, and trained models are available at https://github.com/ tencent-ailab/ICML21\_OAXE. \\
\hline
Open Datasets & True & True & We conducted experiments on major benchmarking datasets that are widely-used in previous NAT studies (Gu et al., 2018; Shao et al., 2020; Ma et al., 2019; Saharia et al., 2020): WMT14 English German (En De, 4.5M sentence pairs), WMT16 English Romanian (En Ro, 0.6M sentence pairs). ... We use the dataset released by Ott et al. (2018) for evaluating translation uncertainty, which consists of ten human translations for 500 sentences taken from the WMT14 En-De test set. \\
\hline
Dataset Splits & True & True & The training set consists of 300K instances, in which the target is an ordering sampled from a given set of ordering modes from a categorical distribution. Both the validation and test sets consist of 3K instances and all the ordering modes serves as the references for the test sets. \\
\hline
Hardware Specification & False & False & The paper mentions that Hungarian Match was implemented with a CPU-version python package and that training speed is 1.36 times slower, but it does not provide specific details about the CPU or any other hardware components like GPU models, memory, or cloud instance types used for experiments. \\
\hline
Software Dependencies & False & False & The paper mentions software like 'python package scipy', 'PyTorch', 'Adam', and 'Fairseq' but does not provide specific version numbers for these dependencies. \\
\hline
Experiment Setup & True & True & We trained batches of approximately 128K tokens using Adam (Kingma \& Ba, 2015). The learning rate warmed up to 5 10 4 in the first 10K steps, and then decayed with the inverse square-root schedule. We trained all models for 300k steps, measured the validation BLEU at the end of each epoch, and averaged the 5 best checkpoints. \\
\hline
\multicolumn{4}{|l|}{\textbf{ICLR 2015 - Move Evaluation in Go Using Deep Convolutional Neural Networks}} \\
\hline
Research Type & Empirical & Empirical & We train a large 12-layer convolutional neural network by supervised learning from a database of human professional games. The network correctly predicts the expert move in 55\% of positions, equalling the accuracy of a 6 dan human player. When the trained convolutional network was used directly to play games of Go, without any search, it beat the traditional-search program Gnu Go in 97\% of games, and matched the performance of a state-of-the-art Monte-Carlo tree search that simulates two million positions per move. \\
\hline
Affiliation & Industry & Industry & *** University of Toronto ***@cs.toronto.edu ***1, ***2, ***1 Google Deep Mind1, Google Brain2 \{***,***,***\}@google.com \\
\hline
Pseudocode & False & False & The paper describes its methods in prose, but does not include any structured pseudocode or algorithm blocks. \\
\hline
Open Source Code & False & False & The paper does not provide any concrete access information (e.g., specific repository link, explicit code release statement, or code in supplementary materials) for the methodology described. \\
\hline
Open Datasets & False & False & The dataset used in this work comes from the KGS Go Server. It consists of sequences of board positions st for complete games played between humans of varying rank. We collected 29.4 million board-state next-move pairs (st, at) corresponding to 160,000 games. While the source is named, no specific link, DOI, formal citation with authors/year, or repository for the collected dataset is provided for public access. \\
\hline
Dataset Splits & True & True & The dataset was split into a training set of 27.4 million board-state next-move pairs and a test set of 2 million. This split was done before shuffling, so this corresponds to a test set with distinct games. \\
\hline
Hardware Specification & True & True & Using a single machine with Intel Xeon CPU E5-2643 v2 @ 3.50GHz and Ge Force GTX Titan Black GPU, we are able to maintain a MCTS search at approximately 47,000 rollouts per second, without dropping CNN evaluations. \\
\hline
Software Dependencies & False & False & The paper mentions methods like 'asynchronous stochastic gradient descent (Dean et al., 2012)' but does not provide specific software names with version numbers (e.g., 'PyTorch 1.9', 'CUDA 11.1') for libraries or frameworks used in the implementation. \\
\hline
Experiment Setup & True & True & We used a deep convolutional neural network with 12 weight matrices for each of 12 layers and rectified linear non-linearities. The first hidden layer s filters were of size 5 5 and the remainder were of size 3 3, with a stride of 1. ... For training the network, we used asynchronous stochastic gradient descent (Dean et al., 2012) with 50 replicas each on its own GPU. All parameters were initialized randomly from a uniform[-0.05, 0.05]. Each replica was trained for 25 epochs with a batchsize of 128, a fixed learning rate of 0.128 normalized by batchsize, and no momentum. \\
\hline
\caption{A sample of LLM-generated reproducibility variable classifications for 10 papers from the evaluation dataset. Papers were selected to span most years and all five conferences covered by the study, and to include at least one misclassification for each reproducibility variable except research type, for which the LLM achieved 100\% accuracy on the evaluation dataset. For each reproducibility variable, the Manual Label column reports the human annotation, the LLM Label column reports the LLM classification, and the LLM Supporting Text column reproduces the text extracted from the paper by the LLM in support of its classification. Author names in the Supporting Text for the affiliation variable have been replaced with ***. Rows where the LLM Label disagrees with the Ground Truth are shown in bold.}
\label{tab:selected-results}
\end{longtable}
 
\end{landscape}

\section{LLM Classification of Negated Variables}
\label{sec:negated-variables}

A common failure mode in keyword-based approaches is misclassifying explicit statements of non-sharing as evidence of open artifacts. 
To illustrate that our LLM-based method has the capability to handle negation, we present five representative examples for the open datasets variable, drawn from the manual evaluation dataset. 
The Label column reports the Boolean value assigned by the LLM: TRUE indicates that the paper uses at least one well-known public dataset or shares a dataset via a URL, DOI, or formal citation; FALSE indicates that no such evidence was found. 
In all cases where a dataset is described as not publicly available, the LLM correctly assigned FALSE. The NeurIPS 2022 example is labeled TRUE because the paper uses a publicly available benchmark (WIKITEXT103) despite also relying on a non-public dataset, demonstrating that the model does not require all datasets in a paper to be public to assign TRUE.

\newgeometry{left=0.75in,right=0.75in,top=1in,bottom=1in}
\begin{landscape}

\setlength{\LTcapwidth}{\linewidth}
\begin{longtable}{|p{0.1\linewidth}|p{0.25\linewidth}|p{0.5\linewidth}|p{0.1\linewidth}|}
\hline
\textbf{Conference} & \textbf{Paper Title} & \textbf{Paper Text} & \textbf{Open Dataset LLM Label} \\
\hline
\endfirsthead
\hline
\textbf{Conference} & \textbf{Paper Title} & \textbf{Paper Text} & \textbf{Open Dataset LLM Label} \\
\hline
\endhead
\hline
\endfoot
NeurIPS 2019 & No Press Diplomacy: Modeling Multi-Agent Gameplay & We are going to release the dataset along with the game engine\textsuperscript{5}. Detailed dataset statistics are shown in Table 1. \newline \textsuperscript{5}Researchers can request access to the dataset by contacting webdipmod@gmail.com. An executive summary describing the research purpose and execution of a confidentiality agreement are required. & False \\
\hline
NeurIPS 2022 & Memorization Without Overfitting: Analyzing the Training Dynamics of Large Language Models & Similarly, while for most of our experiments we section § A.4. use WIKITEXT103 benchmark which is publicly available, some of our experiments run on the ROBERTA dataset which is not publicly available, and therefore, we are unable to release the exact data to re-create those experiments. & True \\
\hline
NeurIPS 2023 & GlucoSynth: Generating Differentially-Private Synthetic Glucose Traces & Limitations. In order to train on a huge set of glucose traces, we used a private dataset, not publicly available (one of the motivations for this project was actually to share a synthetic version of these traces). That being said, smaller samples of glucose traces with similar patient populations are available at OpenHumans [30] and T1D Exchange Registry [31]. In addition, one of the reasons our privacy results perform well is because we use two separate datasets for the training of the motif causality block and the GAN. However, this may be a limiting factor for others that do not have a large enough set of traces available to be able to train adequately on partitioned data. & False \\
\hline
NeurIPS 2024 & Covariate Shift Corrected Conditional Randomization Test & Replication code for our simulation studies is submitted as supplementary
material. It will also be made publicly available on GitHub once our paper is accepted. The COVID data set used for the real example in our paper is not publicly available due to privacy constraints. & False \\
\hline
NeurIPS 2024 & Learning Social Welfare Functions & The dataset we rely on (which is not publicly available) comes from the work of Lee et al. [11] with a US-based nonprofit that operates an on-demand donation transportation service supported by volunteers. & False \\
\hline
\caption{Five examples of LLM classification for the open datasets variable, drawn from the manual evaluation dataset, illustrating the model's handling of negation and partial availability..}
\label{tab:negated-variables}
\end{longtable}
 
\end{landscape}

\end{document}